\documentclass[11pt]{article}

\usepackage[final]{acl}

\usepackage{times}
\usepackage{latexsym}
\usepackage[most]{tcolorbox} 
\usepackage{listings}        
\tcbuselibrary{breakable}

\usepackage[T1]{fontenc}

\usepackage[utf8]{inputenc}

\usepackage{microtype}

\usepackage{inconsolata}

\usepackage{graphicx}
\usepackage{amsmath}
\usepackage{amssymb}
\usepackage{booktabs}
\usepackage{multirow}
\usepackage{multicol}
\usepackage{subcaption}
\usepackage{subcaption}
\usepackage{url}
\usepackage{booktabs}

\usepackage{algorithmic}
\usepackage{algorithm}

\usepackage{booktabs}
\usepackage{tcolorbox}
\tcbuselibrary{breakable}  

\usepackage{xcolor}
\usepackage{tcolorbox}
\tcbuselibrary{skins,breakable}

\title{Cognitive-Uncertainty Guided Knowledge Distillation for Accurate Classification of Student Misconceptions}

\author{
  Qirui Liu$^{1,2,*}$ \quad
  Hao Chen$^{2,*}$ \quad
  Weijie Shi$^{3}$ \quad
  Jiajie Xu$^{4}$ \quad
  Jia Zhu$^{5,\dagger}$ \\\\
  $^{1}$South China University of Technology,
  $^{2}$Tencent Financial Technology \\
  $^{3}$The Hong Kong University of Science and Technology,
  $^{4}$Soochow University \\
  $^{5}$Zhejiang Key Laboratory of Intelligent Education Technology and Application, \\
  Zhejiang Normal University
}

\begin{document}
\maketitle

\let\thefootnote\relax
\footnotetext{$^{*}$\,Equal contribution.}
\footnotetext{$^{\dagger}$\,Corresponding Author.}

\begin{abstract}

Accurately identifying student misconceptions is crucial for personalized education but faces three challenges: (1) data scarcity with long-tail distribution, where authentic student reasoning is difficult to synthesize; (2) fuzzy boundaries between error categories with high annotation noise; (3) deployment paradox—large models overlook unconventional approaches due to pretraining bias and cannot be deployed on edge, while small models overfit to noise.
Unlike traditional methods that increase diversity through large-scale data synthesis, 
we propose a two-stage knowledge distillation framework that mines high-value samples from existing data. The first stage performs standard distillation to transfer task capabilities. The second stage introduces a dual-layer marginal selection mechanism based on cognitive uncertainty, identifying four types of critical samples based on teacher model uncertainty and confidence differences. For different data subsets, we design difficulty-adaptive mechanism to balance hard/soft label contributions, enabling student models to inherit inter-class relationships from teacher soft labels while distinguishing ambiguous error types.
Experiments show that with augmented training on only 10.30\% of filtered samples, we achieve MAP@3 of 0.9585 (+17.8\%) on the MAP-Charting dataset, and using only a 4B parameter model, we attain 84.38\% accuracy on cross-topic tests of middle school algebra misconception benchmarks, significantly outperforming sota LLM (67.73\%) and standard fine-tuned 72B models (81.25\%).
Our code is available at \url{https://github.com/RoschildRui/acl2026_map}.


\end{abstract}

\section{Introduction}

Understanding how students think and solve problems remains a core challenge in educational research~\cite{HIT,chanllenge-1,chanllenge-2}. Traditional assessment methods focusing solely on answer correctness overlook students' reasoning processes, failing to reveal cognitive obstacles or recognize partially valid thinking within incorrect answers. This limits teachers' ability to provide personalized feedback and prevents platforms from making adaptive adjustments based on actual thinking trajectories~\cite{AutoTutor,p-VAE}.

\begin{figure*}
\centering
\includegraphics[width=0.95\linewidth]{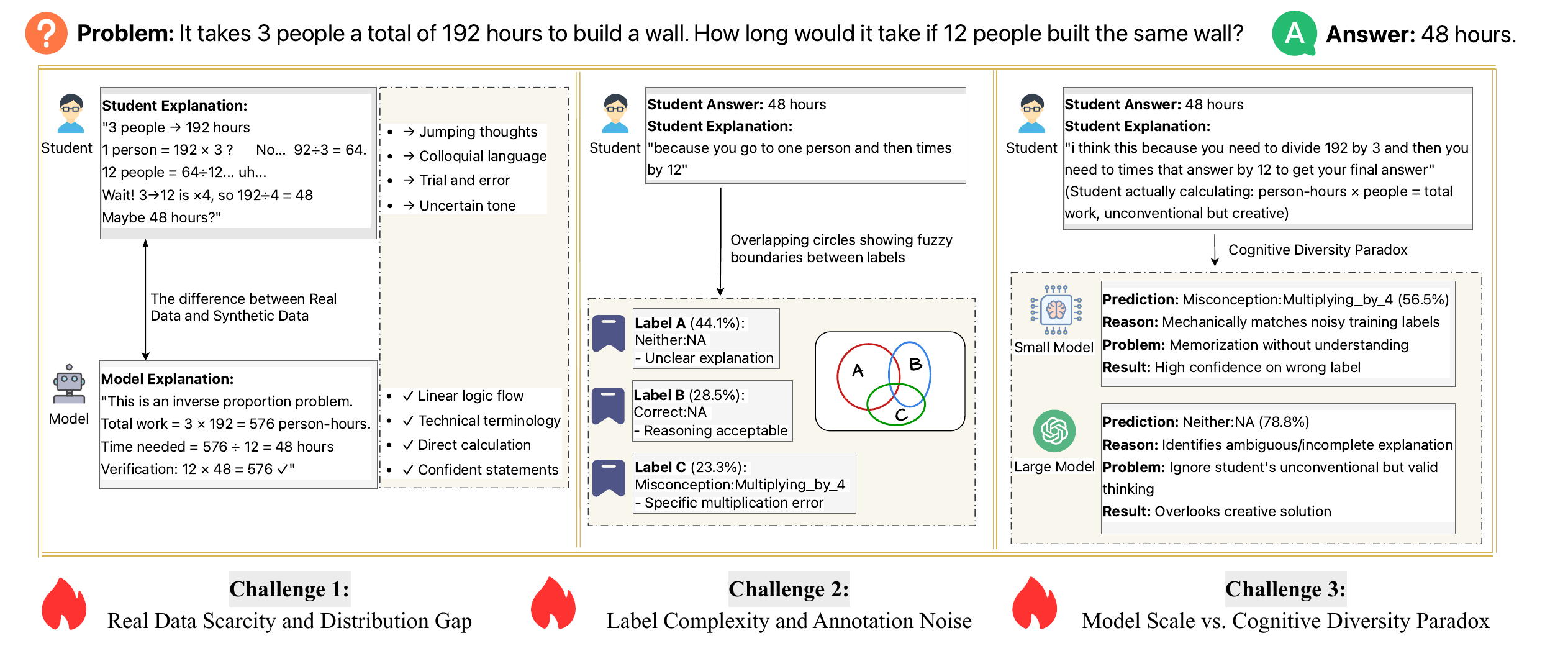}
\caption{The overview of the key points and corresponding challenges.}
\label{fig:intro}
\end{figure*}

This limitation is particularly acute in mathematics education, where identical incorrect answers may stem from completely different reasoning paths revealing distinct conceptual misunderstandings~\cite{MaE,misconception-1}. For example, solving $2x + 3 = 11$, one student might correctly subtract 3 then divide by 2 but make a calculation error, while another might directly divide 11 by 2 due to conceptual confusion. Both produce wrong answers requiring fundamentally different interventions~\cite{Bridg-gap}. This necessitates shifting from evaluating "answer correctness" to understanding "problem-solving approaches," which recent NLP advances now make feasible through automatic classification of student reasoning processes~\cite{MathEDU}.

However, accurate classification faces three core challenges:

\textbf{Challenge 1: Data scarcity and distribution gap.} Real student process data is severely limited with long-tail distribution~\cite{MathEDU}. While existing work uses LLMs to generate synthetic data~\cite{data-survey}, real student reasoning features colloquial language, reasoning jumps, and logical errors that models cannot accurately replicate. LLM-generated text's superior logic and fluency creates significant distribution gaps, limiting generalization to real scenarios.

\textbf{Challenge 2: Label complexity and annotation noise.} Student misconception labels are numerous with fuzzy boundaries between error types and even correct/incorrect categories~\cite{Bridg-gap}, causing substantial annotation noise. Traditional hard-label models cannot learn subtle inter-category differences or handle inherent uncertainty, performing poorly on complex ambiguous errors.

\textbf{Challenge 3: Model scale and cognitive diversity paradox.} Students' diverse thinking often produces unconventional but reasonable solutions within their cognitive framework~\cite{MaE}. Small models meet deployment needs but overfit to noisy labels; large models possess rich knowledge but systematically overlook non-standard approaches due to pretraining bias~\cite{gsm-ic}, force-fitting innovative solutions into existing frameworks and misjudging error types. Educational privacy requirements and edge device limitations further prevent direct large model deployment.

Addressing these challenges, unlike traditional methods that rely on large-scale data synthesis~\cite{data-survey}, we propose a two-stage knowledge distillation framework that strategically mines high-value samples from existing data. The first stage performs standard distillation for basic task capability transfer; the second stage introduces a dual-layer marginal selection mechanism based on cognitive uncertainty, identifying Near-miss (correct but uncertain) and Hard-hard (severely incorrect) critical samples through teacher uncertainty and confidence differences, with difficulty-adaptive loss functions dynamically balancing hard/soft labels to help students distinguish complex error types while inheriting inter-class relationships. This approach produces lightweight models that accurately identify diverse student approaches while meeting practical requirements for privacy protection and edge deployment.

Our main contributions include:
\begin{itemize}
\item A dual-layer marginal selection mechanism based on cognitive uncertainty that precisely filters high-value real samples for incremental training, improving limited-data performance without synthetic data dependency.
\item Difficulty-adaptive loss functions with dynamic soft/hard label weighting based on sample difficulty, addressing label noise and boundary ambiguity while improving discrimination of confusable categories.
\item Addressing large models' diversity oversight from pretraining bias through two-stage distillation, balancing knowledge transfer with cognitive openness to maintain inclusiveness toward non-standard solutions.
\end{itemize}
\section{Related Work}

\begin{figure*}
    \centering
    \includegraphics[width=0.95\linewidth]{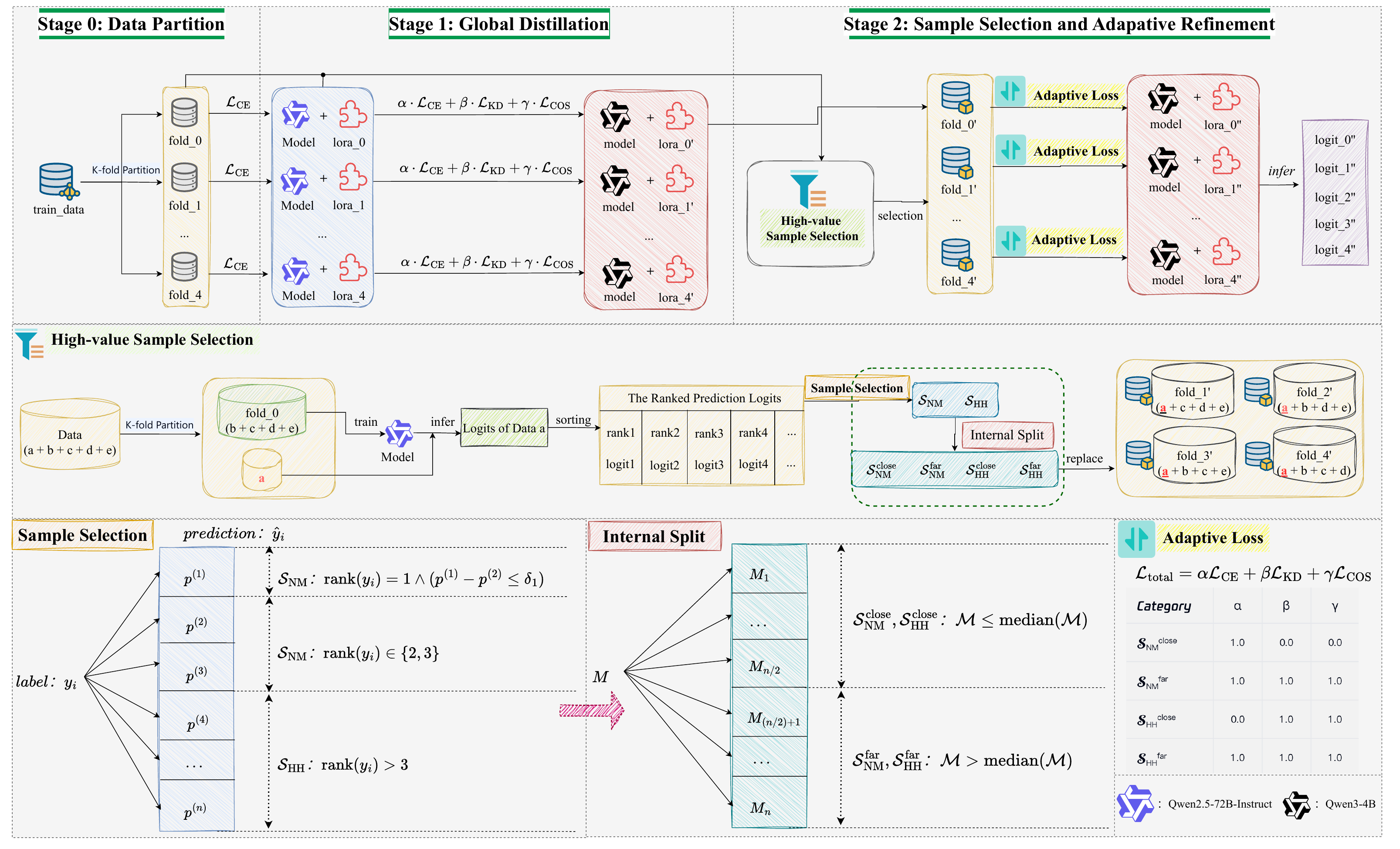}
    \caption{Overview of our two-stage distillation framework, consisting of data partitioning, two-stage knowledge distillation, and sample selection.}
    \label{fig:pipeline}
\end{figure*}

In this section, we first introduce the evolution in the education domain from answer scoring to process understanding, and then discuss related techniques for handling data scarcity and label noise.

\subsection{Student Reasoning Assessment and Misconception Diagnosis}

Educational assessment research has undergone a paradigm shift from result-oriented evaluation to process understanding. Early studies focused primarily on the automatic determination of answer correctness \cite{chanllenge-1,chanllenge-2}, achieving scoring accuracy close to human level, but lacking the ability to diagnose the underlying causes of errors. Recent work has begun to explore misconception identification. For example, \citet{MaE,MathEDU} proposed fine-grained error classification frameworks in the algebra domain, and \citet{Bridg-gap} summarized 55 types of algebraic misconceptions and built diagnostic benchmarks. However, these approaches still rely mainly on the final answer rather than the complete reasoning process.

Directly analyzing students’ reasoning processes is key to understanding cognitive barriers. \citet{gsm-ic} found that even advanced language models can fail in reasoning when confronted with explanations containing irrelevant information or logical leaps, a phenomenon that mirrors characteristics commonly observed in real student responses. However, obtaining high-quality student process data is extremely challenging. Although large language models (LLMs) can generate synthetic data \cite{data-survey}, such data tends to exhibit overly standardized expressions that differ markedly from the colloquial, non-standard reasoning of real students, resulting in poor generalization for models trained solely on synthetic data.

To address the distribution bias in synthetic data, we propose a sample selection strategy based on cognitive uncertainty. This approach accurately identifies the most critical \emph{near-miss} and \emph{hard-hard} samples from limited real data, which are most influential to the decision boundary, thereby avoiding reliance on synthetic data.

\subsection{Sample-Efficient Learning under Label Noise}

In the presence of data scarcity and label noise, the machine learning community has developed various strategies. Curriculum learning \cite{curriculum,self-evolving-curriculumllmreasoning} improves learning efficiency by arranging training in an easy-to-hard sequence; subsequent work further introduced uncertainty-based active sample selection \cite{activelearning-1,activelearning-2,activelearning-3}, prioritizing the learning of samples with the highest information gain. The core idea behind these methods is to identify and focus on key samples that can significantly improve the decision boundary~\cite{dis-aug-1,dis-aug-2,dis-aug-3}.

Knowledge distillation \cite{distillation,distillation-survey} offers another path for handling label noise. The soft labels from teacher models embed rich inter-class relational information, which can mitigate the impact of noisy hard labels. Recent adaptive distillation methods \cite{adaptive-distillation,spot-adaptive-distillation,hard-adaptive-distillation} further propose dynamically adjusting the weights of soft and hard labels based on sample difficulty, enabling more fine-grained knowledge transfer~\cite{distill-sup-0,distill-sup-1,distill-sup-2}. However, existing approaches face unique challenges in educational scenarios: biases from large model pretraining make it difficult to accept non-standard student reasoning, and direct distillation can cause small models to excessively inherit these biases~\cite{dis-sup-3,dis-sup-4,dis-sup-5,dis-sup-6}.

To address the cognitive bias issue in large models, we propose a difficulty-adaptive loss function that dynamically adjusts the weighting of soft and hard labels, allowing small models to inherit knowledge from large models while maintaining tolerance for non-standard student expressions.

\section{Methodology}

To address the dual challenges of data and labels in identifying students' problem-solving approaches, we propose an Adaptive Knowledge Distillation Framework for High-Value Samples. Instead of relying on synthetic data with distribution gaps, our framework accurately identifies and exploits the most valuable samples from limited authentic data through a two-stage training process of sample selection and adaptive learning, guiding models to tackle real-world ambiguity and complexity.

\subsection{Problem Formulation}



Our goal is to train a classification model $f_s$ that accurately assigns student math problem-solving processes $x_i$ (including problem description, answer, correctness, and metadata) to predefined reasoning labels $y_i$. The label set $\mathcal{Y}$ contains "Correct", multiple misconception categories ($\text{Misconception}_1, ..., \text{Misconception}_K$), and "Neither" for unclassifiable cases.

Given training set $\mathcal{D} = \{(x_i, y_i)\}_{i=1}^N$ with $N$ real student samples, we optimize student model $f_s$ parameters using pretrained teacher models $f_t$ to address label noise and data sparsity. The model outputs probability distribution $p(y|x) = \text{softmax}(z/\tau)$, where $z \in \mathbb{R}^{|\mathcal{Y}|}$ represents logits and $\tau$ is the temperature coefficient for distribution smoothing in knowledge distillation.

\subsection{Stage One: Global Knowledge Distillation and Preliminary Learning}

In student thinking discrimination, we face a fundamental contradiction: large models possess abundant knowledge but ignore thinking diversity due to overconfidence in prior distributions; small models are flexible but lack necessary knowledge foundations. Stage one thus uses knowledge distillation to equip small models with basic knowledge structures while maintaining openness to non-standard ideas.

We employ $n$-fold cross-validation to generate soft labels and prevent overfitting. Using \texttt{StratifiedKFold}~\cite{k-fold} based on label distribution, we split data into $n$ subsets. For each fold $k \in \{1, ..., n\}$, train teacher model $f_t^{(k)}$ on $\mathcal{D} \setminus \mathcal{D}_k$ and generate soft labels $\mathbf{y}_{\text{soft}}^{(i)}$ for samples in $\mathcal{D}_k$.

We then train $n$ student models with a loss function combining three objectives:
\begin{multline}
\mathcal{L} = \alpha \cdot \mathcal{L}_{\mathrm{CE}}(f_s(x_i), y_i) 
+ \beta \cdot \mathcal{L}_{\mathrm{KD}}(f_s(x_i), \mathbf{y}_{\text{soft}}) \\
+ \gamma \cdot \mathcal{L}_{\mathrm{COS}}(f_s(x_i), \mathbf{y}_{\text{soft}})
\end{multline}
where $\mathcal{L}_{\mathrm{CE}}$ ensures basic classification accuracy, $\mathcal{L}_{\mathrm{KD}}$ transfers inter-class relationships, and $\mathcal{L}_{\mathrm{COS}}$ constrains student-teacher consistency in representation space.

To ensure the stability and cross-model generalization of hyperparameters, we conducted comprehensive grid search and ablation experiments on the loss weights ($\alpha, \beta, \gamma$) for Stage 1 (see Appendix \ref{app:hparam_generalization}). 
\subsection{Stage Two: High-Value Sample Selection and Adaptive Refinement}
After initial training, we employ a two-tier margin selection mechanism to identify the most valuable samples for model enhancement, then perform targeted training on this subset using difficulty-adaptive loss functions.

\subsubsection{Two-Tier Margin-Based Sample Selection}
In our framework, the teacher's cognitive uncertainty serves only as a guiding signal for high-value sample selection rather than a hard constraint (see Appendix \ref{app:teacher_uncertainty}).

This mechanism extracts samples that most effectively reveal model weaknesses and provide maximum informational value, directly addressing data scarcity and distribution discrepancies through deep exploration of existing data rather than blind expansion.

\textbf{Uncertainty-Based Difficulty Partition}  
The most beneficial samples are not ones already mastered, but those near decision boundaries or beyond current understanding. Based on teacher model $f_t$ predictions from stage one, we identify two high-value types:

\begin{itemize}
    \item \textbf{Near-miss samples ($\mathcal{S}_{\text{NM}}$):} Samples with correct but low-confidence predictions, or incorrect predictions where the correct answer is almost reached. These boundary-adjacent samples are crucial for fine-grained discrimination. Formally:

\begin{equation}
    \begin{split}
    \mathcal{S}_{\mathrm{NM}} = \{(x_i, y_i) : &[(\hat{y}_i = y_i) \land (p^{(1)} - p^{(2)}) \\ \leq \delta)]
    & \lor \ \text{rank}(y_i) \in \{2, 3\} \}
    \end{split}
\end{equation}
Here, $p^{(k)}$ represents the $k$-th highest predicted probability from the model, $\hat{y}_i$ denotes the predicted class, $\text{rank}(y_i)$ is the position of the true label $y_i$ in the sorted prediction probabilities, and $\delta$ is a small confidence margin threshold (e.g., 0.05).

\item \textbf{Hard-hard samples ($\mathcal{S}_{\text{HH}}$):} These are samples for which the model’s predictions are “grossly incorrect,” meaning the predicted probability ranking of the true label is very low. Such samples expose fundamental knowledge gaps or severe misunderstandings of certain complex concepts.
    \begin{equation}
        \begin{split}
        \mathcal{S}_{\text{HH}} = \{(x_i, y_i) :~ &\text{rank}(y_i) > 3 \} \\
        \end{split}
    \end{equation}
\end{itemize}

Through this tiered partitioning, we narrow the training focus from the entire dataset to $\mathcal{D}_{\text{selected}} = \mathcal{S}_{\text{NM}} \cup \mathcal{S}_{\text{HH}}$, achieving the first stage of concentration on high-value samples.

\textbf{Fine-Grained Differentiation Based on Probability Margin}
Even within $\mathcal{S}_{\text{NM}}$ and $\mathcal{S}_{\text{HH}}$, there exists significant variation in sample difficulty.  
To enable more fine-grained adaptive learning, we introduce a composite difficulty measure that combines the prediction probability margin with distributional uncertainty.  
The composite difficulty metric ${M}(x_i, y_i)$ is defined as follows:

1. \textbf{Probability Margin:}
\begin{equation}
d(x_i, y_i) = |p_s(y_i|x_i) - \max_{j \in \mathcal{Y}} p_s(j|x_i)|
\end{equation}
This reflects the model’s direct fitting degree to the ground truth label.  
A larger $d$ indicates that the model's understanding deviates more from the true label;  
a smaller $d$ suggests that the model is closer to correctly understanding it.

2. \textbf{Prediction Entropy:}
\begin{equation}
H(x_i) = -\sum_{j \in \mathcal{Y}} p_s(j|x_i) \log p_s(j|x_i)
\end{equation}
This reflects the overall dispersion of the model’s predictions.

3. \textbf{Composite Difficulty Metric:}
\begin{equation}
\mathcal{M}(x_i, y_i) = d(x_i, y_i) \cdot e^{-H(x_i)}
\end{equation}
Here, the margin $d(x_i, y_i)$ represents the prediction bias with respect to the ground truth;  
the entropy $H(x_i)$ adjusts the weight of the bias, with $e^{-H(x_i)}$ amplifying difficulty when entropy is low, and attenuating it when entropy is high.  
This metric is designed to capture two complementary dimensions of difficulty.

Based on $M$, we further divide $\mathcal{S}_{\text{NM}}$ and $\mathcal{S}_{\text{HH}}$ into “close” and “far” subsets according to the median:



\begin{equation}
\mathcal{S}_{t}^{\text{close}} = \{(x_i, y_i) \in \mathcal{S}_{t} : \mathcal{M}(x_i, y_i) \leq \text{median}(\mathcal{M}_{t})\}
\end{equation}
\begin{equation}
\mathcal{S}_{t}^{\text{far}} = \{(x_i, y_i) \in \mathcal{S}_{t} : \mathcal{M}(x_i, y_i) > \text{median}(\mathcal{M}_{t})\}
\end{equation}
where $t \in \{\text{NM}, \text{HH}\}$.

This dual-dimensional characterization constructed from the probability margin and prediction entropy precisely depicts different levels of sample difficulty, providing a reliable basis for subsequent adaptive loss design.
Hyperparameter search confirms $\delta=0.05$ and $K=5$ as optimal settings (see Appendix \ref{app:hparam_generalization}).

\subsubsection{Difficulty-Adaptive Loss Function}

To address the issues of label complexity and annotation noise, we design an adaptive loss function that dynamically fuses information from hard labels and soft labels, with its weights adjusted according to the sample categories identified in the previous section. The complete definition of the total loss is:

\begin{equation}
\mathcal{L}_{\text{total}}
= \alpha \mathcal{L}_{\text{CE}} 
+ \beta \mathcal{L}_{\text{KD}}
+ \gamma \mathcal{L}_{\text{COS}}
\end{equation}
where $\mathcal{L}_{\text{CE}}$ is the standard cross-entropy loss based on the ground-truth (hard) label $y_i$; $\mathcal{L}_{\text{KD}}$ is the knowledge distillation loss that guides the student model $f_s$ to mimic the soft probability distribution output by the teacher model $f_t$; and $\mathcal{L}_{\text{COS}}$ is the cosine embedding loss, which constrains the directional consistency between the student and teacher models in the representation space. The specific formulations are defined as:

\begin{equation}
\mathcal{L}_{\text{CE}} = -\log p_s(y_i|x_i)
\end{equation}
\begin{equation}
\mathcal{L}_{\text{KD}} = \tau^2 \cdot \text{KL}(p_t(\cdot|x_i) \| p_s(\cdot|x_i))
\end{equation}
\begin{equation}
\mathcal{L}_{\text{COS}} = 1 - 
\cos\big(p_s(\cdot \mid x_i),\, p_t(\cdot \mid x_i)\big) 
\end{equation}

The key lies in the adaptive allocation strategy of the coefficients $(\alpha, \beta, \gamma)$ based on different sample categories. For samples with high uncertainty near the decision boundary, strong constraints from hard labels are necessary to avoid the soft-label smoothing effect. Samples near but slightly away from the boundary benefit from both the precision of hard labels and the inter-class relationships of soft labels. When predictions are close to ground-truth but exhibit significant deviations, soft labels are favored to mitigate noise impact. For extremely difficult samples, both hard- and soft-label guidance need to be strengthened. 
The specific weight allocation for each sample category is detailed in Appendix~\ref{sec:loss_weights}.
We also visualize sample characteristics and prediction differences to verify selection effectiveness (see Appendix \ref{app:sample_visualization}).

\begin{table*}[t]
\centering
\caption{Performance comparison on two benchmark datasets. The best results in each category are shown in \textbf{bold}.}
\label{tab:combined_results}
\resizebox{\textwidth}{!}{%
\begin{tabular}{l|cccc|cccc}
\toprule
\multirow{2}{*}{\textbf{Method}} & \multicolumn{4}{c|}{\textbf{MAP-Charting}} & \multicolumn{4}{c}{\textbf{Algebra Misconception Benchmark}} \\
\cmidrule{2-9}
 & \textbf{MAP@3} & \textbf{MAP@10} & \textbf{Accuracy} & \textbf{F1@3} & \textbf{MAP@3} & \textbf{MAP@10} & \textbf{Accuracy} & \textbf{F1@3} \\
\midrule
\multicolumn{9}{l}{\textit{Prompting-based Methods}} \\
GPT-5 & \textbf{0.8137} & \textbf{0.8145} & \textbf{0.7225} & \textbf{0.4626} & \textbf{0.7409} & \textbf{0.7418} & \textbf{0.6773} & \textbf{0.4091} \\
Claude-4-Sonnet & 0.7833 & 0.7841 & 0.6914 & 0.4579 & 0.6636 & 0.6645 & 0.5636 & 0.3932 \\
DeepSeek-V3 & 0.7665 & 0.7673 & 0.6601 & 0.4505 & 0.6485 & 0.6494 & 0.5545 & 0.3815 \\
GPT-OSS-120B & 0.7661 & 0.7669 & 0.6794 & 0.4375 & 0.6550 & 0.6559 & 0.5680 & 0.3725 \\
Qwen-2.5-72B (prompting) & 0.7285 & 0.7293 & 0.6222 & 0.4328 & 0.6280 & 0.6289 & 0.5320 & 0.3670 \\
\midrule
\multicolumn{9}{l}{\textit{Fine-tuned-based Methods}} \\
Qwen-2.5-72B (fine-tuned) & \textbf{0.9497} & \textbf{0.9501} & \textbf{0.9014} & \textbf{0.4993} & \textbf{0.8438} & \textbf{0.8612} & \textbf{0.8125} & \textbf{0.4375} \\
Qwen-3-4B (fine-tuned) & 0.9472 & 0.9475 & 0.8987 & 0.4992 & 0.7552 & 0.7669 & 0.7188 & 0.4062 \\
Gemma-2-9B (fine-tuned) & 0.9439 & 0.9442 & 0.8919 & 0.4992 & 0.7708 & 0.7862 & 0.7188 & 0.4219 \\
Llama-3.1-8B (fine-tuned) & 0.9453 & 0.9456 & 0.8954 & 0.4990 & 0.7760 & 0.7917 & 0.7500 & 0.4062 \\ 
\midrule
\multicolumn{9}{l}{\textit{Our Method (Two-Stage Distillation)}} \\
Qwen-3-4B + Ours & \textbf{0.9585} & \textbf{0.9587} & \textbf{0.9198} & \textbf{0.4996} & \textbf{0.8750} & \textbf{0.8915} & \textbf{0.8438} & \textbf{0.4531} \\
Gemma-2-9B + Ours & 0.9560 & 0.9562 & 0.9148 & 0.4995 & 0.8015 & 0.8155 & 0.7656 & 0.4375 \\
Llama-3.1-8B + Ours & 0.9553 & 0.9555 & 0.9134 & 0.4995 & 0.7865 & 0.7995 & 0.7564 & 0.4281 \\
\bottomrule
\end{tabular}
}
\vspace{2pt}
\end{table*}

\section{Experiments and Analysis}

\subsection{Experimental Setup}

\subsubsection{Datasets}

We evaluate the proposed method on two complementary benchmarks, which represent different levels of granularity in student misconception detection:

\textbf{MAP-Charting dataset}~\cite{benchmark1}: This dataset contains real reasoning traces of students in multiple-choice mathematics questions. Each sample includes the problem statement, the student's answer, the correctness label, and—critically—the student's written explanation of their reasoning process. Labels are divided into three categories: Correct, Misconception (with specific misconception types), or Neither (indicating vague or irrelevant thinking). This fine-grained dataset consists of 36,695 samples in total.

\textbf{Algebra Misconceptions Benchmark}~\cite{benchmark2}: This benchmark covers 55 types of algebraic misconceptions validated by 145 peer-reviewed studies. Unlike MAP-Charting, this dataset requires only the student’s final answer (without reasoning traces) to identify the misconception type, making it coarser-grained but easier to collect. We randomly select questions from all available topics for evaluation. This dataset contains a total of 220 samples.

\subsubsection{Implementation Details}

For student models, we use three lightweight architectures: Qwen-3-4B~\cite{qwen3}, Gemma-2-9B~\cite{gemma}, and Llama-3.1-8B~\cite{llama-3.1}, with Qwen-2.5-72B~\cite{qwen2.5} as the teacher. All models are fine-tuned via AdamW (batch size=16, 4 gradient accumulation steps): student learning rate $2 \times 10^{-4}$, teacher learning rate $1 \times 10^{-4}$, distillation temperature $\tau = 1.0$, and confidence threshold $\delta = 0.05$ (from preliminary experiments). For second-stage incremental training, student models use learning rate $1 \times 10^{-6}$ and \texttt{max\_grad\_norm}=4, with other configurations unchanged.

\subsubsection{Baseline Methods}

\textbf{Prompting-based models}: We evaluate various prompting strategies for advanced large language models, including in-context learning and chain-of-thought reasoning. Tested models include GPT-5~\cite{gpt-5}, Claude-4-Sonnet~\cite{claude}, DeepSeek-V3~\cite{DeepSeek-V3}, and Qwen-2.5-72B.

\textbf{Classification-based models}: We attach a classification head and fine-tune different sizes of language models, including student models (Qwen-3-4B, Gemma-2-9B, Llama-3.1-8B) and the teacher model (Qwen-2.5-72B) directly fine-tuned without distillation.

\subsubsection{Evaluation Metrics}

We adopt four evaluation metrics to comprehensively assess ranking quality from multiple perspectives: MAP@3, MAP@10, Accuracy, and F1@3. MAP@3 mainly measures the model's ability to identify multiple possible correct answers within the top 3 predictions; MAP@10 extends the scope to the top 10 predictions, evaluating ranking performance on a broader scale; Accuracy measures the overall correctness of predictions; F1@3 considers both precision and recall within the top 3 predictions, reflecting the model’s balanced performance in multi-candidate scenarios. These four metrics allow us to compare model performance in terms of local precision, overall accuracy, and ranking coverage from multiple angles.

\subsection{Main Results}



Table~\ref{tab:combined_results} shows that two-stage distillation markedly boosts MAP@3 and MAP@10 on both MAP-Charting and Algebra Misconception benchmarks, outperforming prompt-based reasoning and direct fine-tuning. In MAP-Charting, the best prompt-based GPT‑5 score (0.8137/0.8145) rises to 0.9497/0.9501 via direct fine-tuning of Qwen-2.5-72B, while the distilled Qwen-3-4B attains 0.9585/0.9587, about 0.9\% above the teacher model and surpassing the 72B model; Gemma-2-9B and Llama-3.1-8B show similar gains. On Algebra Misconception, Qwen-3-4B improves from 0.7552/0.7669 to 0.8750/0.8915 (15.8\%/16.2\%), exceeding GPT‑5 by 18.1\%/20.2\%. Consistent gains across models and metrics indicate stable advantages. Overall, two-stage distillation enables lightweight students to close and sometimes surpass the gap with large teacher models, combining stability, generality, and efficiency.
The 4B-parameter student model outperforms the 72B-parameter teacher model, with key reasons detailed in Appendix \ref{app:student_beats_teacher}.

\subsection{Ablation Study}

Our ablation study consists of two parts.  
The first part removes (ablates) the main components of the method one by one to verify the impact of different modules on model performance.  
The second part takes into account the multi-stage distillation training nature of our method and designs a cross-stage performance comparison experiment.

\subsubsection{Ablation of Main Components}




Table~\ref{tab:ablation_study} reports ablation results for MAP@10, MAP@3, and Accuracy using the two-stage distillation scheme based on Qwen-3-4B. Removing any core component consistently lowers both MAP@10 and MAP@3, indicating stable gains across different evaluation dimensions. On the MAP-Charting dataset, the full method achieves 0.9587/0.9585 (MAP@10/MAP@3) and 0.9198 (Accuracy); removing adaptive loss or high-value sample selection reduces performance by 0.4\%--0.7\%, while omitting second-stage distillation causes the largest drop, with Accuracy falling by over 1.7\%. On the Algebra Misconception benchmark, the full method obtains 0.8915/0.8750 and 0.8438; removing second-stage distillation reduces MAP by more than 10\% and Accuracy to 0.7577. Overall, each stage of the distillation process provides significant, consistent improvements across different evaluation dimensions.

\subsubsection{Performance Across Different Stages}

\begin{figure}[htbp]
    \centering
    \begin{subfigure}{0.48\columnwidth}
        \centering
        \includegraphics[width=\linewidth]{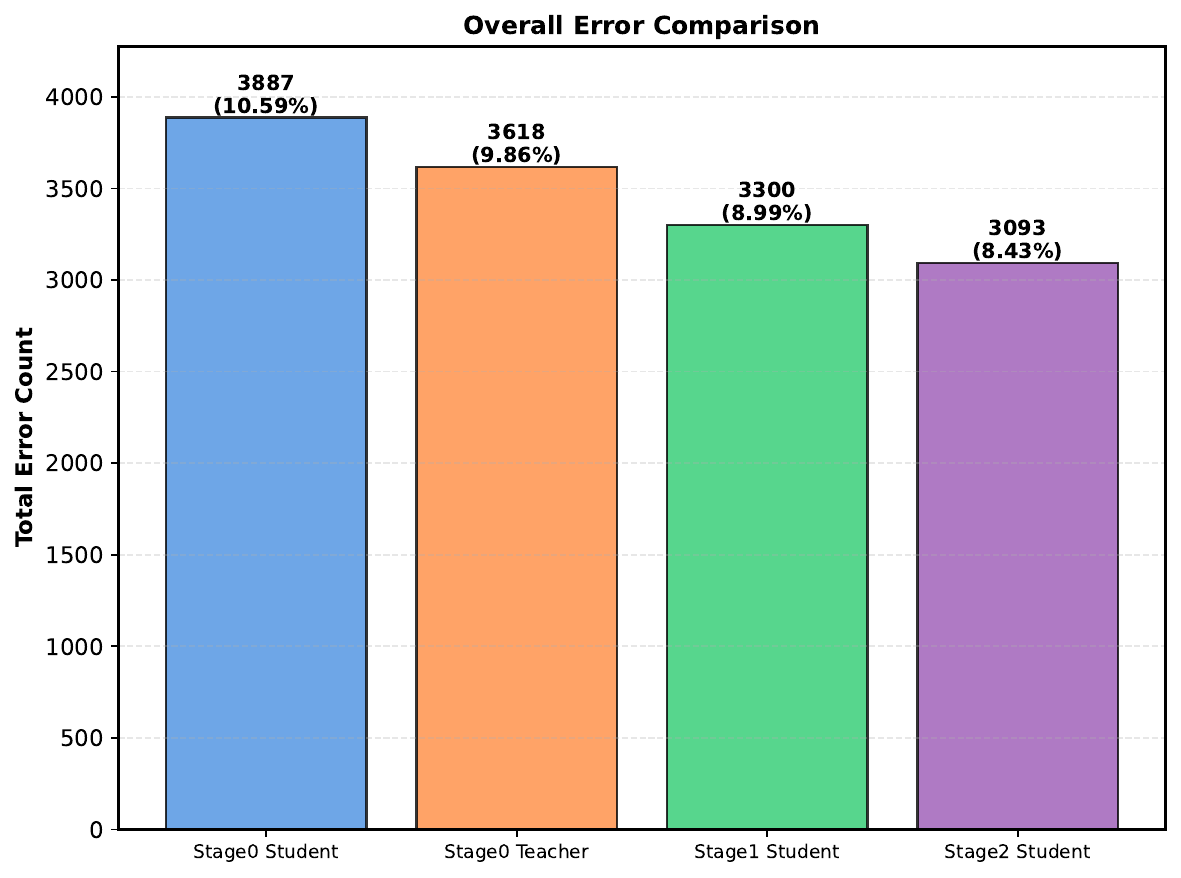}
        \caption{Error count changes across different stage models}
        \label{fig:stage-errors}
    \end{subfigure}
    \hfill
    \begin{subfigure}{0.48\columnwidth}
        \centering
        \includegraphics[width=\linewidth]{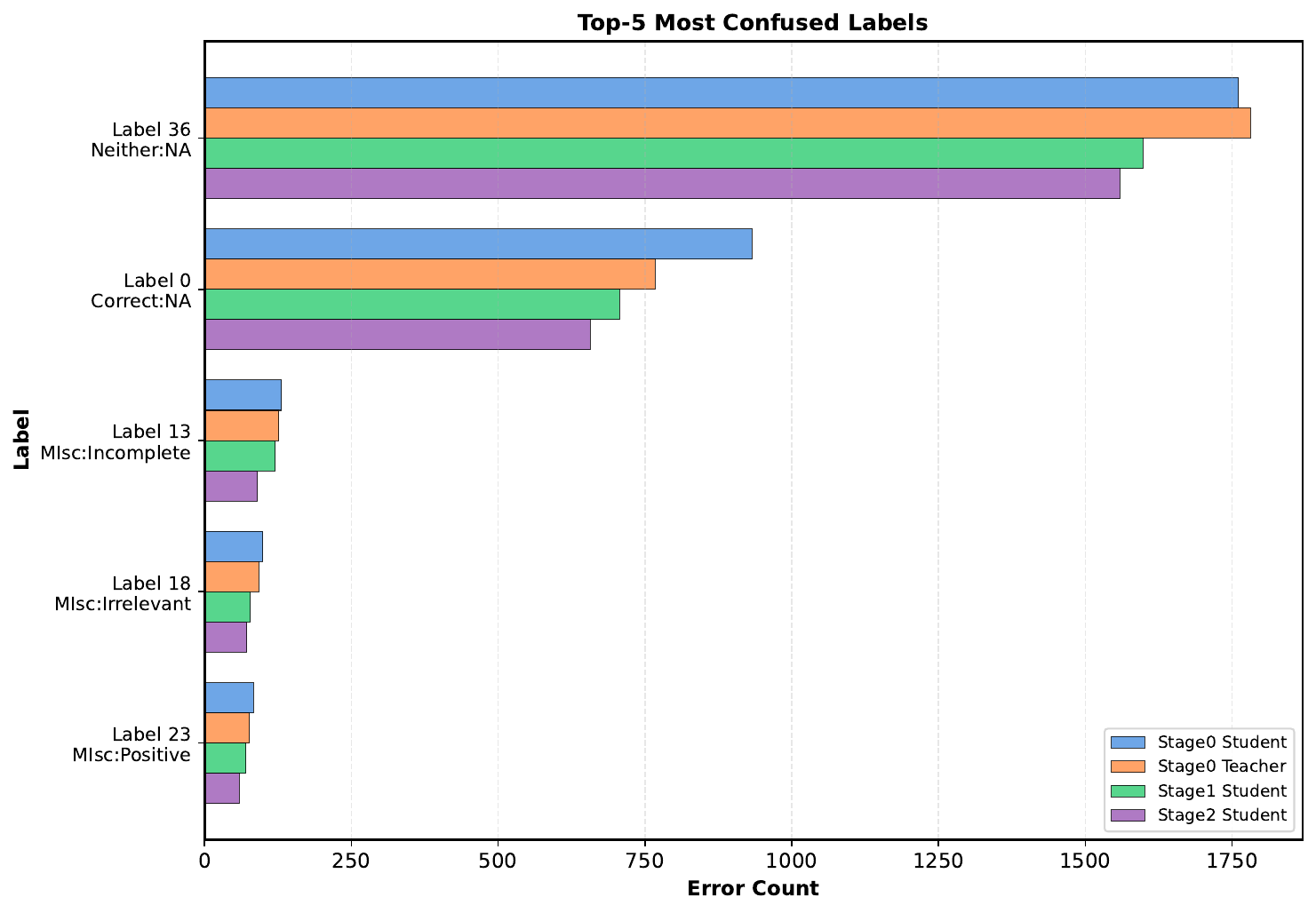}
        \caption{Error count changes in top-5 confused categories}
        \label{fig:class-errors}
    \end{subfigure}

    \caption{Visualization of error counts in multi-stage distillation training.}
    \label{fig:errors}
\end{figure}

Figures~\ref{fig:stage-errors} and~\ref{fig:class-errors} show error count changes across key stages and top-5 confused categories, with later-stage models consistently reducing total and per-category errors. This confirms multi-stage distillation lowers error rates and enhances fine-grained classification, especially for difficult categories. Detailed top-10 and 37-category analyses are in Appendix~\ref{sec:error-analysis-appendix}.

\begin{table}[t]
\centering
\footnotesize
\setlength{\tabcolsep}{3pt}
\caption{Ablation study results. The best result for each benchmark dataset is in \textbf{bold}.}
\label{tab:ablation_study}
\begin{tabular}{l|ccc}
\toprule
\multirow{2}{*}{\textbf{Method Variant}} & \multicolumn{3}{c}{\textbf{MAP-Charting}} \\
\cmidrule{2-4}
 & \textbf{MAP@10} & \textbf{MAP@3} & \textbf{Accuracy} \\
\midrule
Full Method & \textbf{0.9587} & \textbf{0.9585} & \textbf{0.9198} \\
w/o Adaptive Loss & 0.9542 & 0.9540 & 0.9123 \\
w/o Sample Selection & 0.9521 & 0.9519 & 0.9085 \\
w/o Stage-1 Distillation & 0.9548 & 0.9546 & 0.9132 \\
w/o Stage-2 Distillation & 0.9495 & 0.9493 & 0.9024 \\
\midrule
\multirow{2}{*}{\textbf{Method Variant}} & \multicolumn{3}{c}{\textbf{Algebra Misconception}} \\
\cmidrule{2-4}
 & \textbf{MAP@10} & \textbf{MAP@3} & \textbf{Accuracy} \\
\midrule
Full Method & \textbf{0.8915} & \textbf{0.8750} & \textbf{0.8438} \\
w/o Adaptive Loss & 0.8802 & 0.8657 & 0.8321 \\
w/o Sample Selection & 0.8741 & 0.8603 & 0.8269 \\
w/o Stage-1 Distillation & 0.8823 & 0.8679 & 0.8342 \\
w/o Stage-2 Distillation  & 0.8001 & 0.7893 & 0.7577 \\
\bottomrule
\end{tabular}
\end{table}

\subsection{Efficiency Analysis}

To evaluate practical value, we tested inference efficiency on 7,339 samples (Table~\ref{tab:efficiency}). Qwen-3-4B outperforms GPT-5 by 18.0 pp in MAP@3 (0.9599 vs 0.8137) with 187.5$\times$ speedup (0.008\,h vs 1.50\,h), slightly outperforms teacher Qwen-2.5-72B (0.9599 vs 0.9497) with 23.25$\times$ speedup, and surpasses GPT-OSS-120B by 25.3 pp with 137.5$\times$ speedup (runnable on PC). These results confirm our method’s gains in producing accurate, lightweight models for misconception classification.

\begin{table}[t]
\centering
\footnotesize
\setlength{\tabcolsep}{1pt}
\caption{Inference efficiency comparison over 7,339 samples}
\label{tab:efficiency}
\begin{tabular}{lccc}
\toprule
\textbf{Model} & \textbf{MAP@3} & \textbf{Time (h)} & \textbf{Hardware} \\
\midrule
\multicolumn{4}{l}{\textit{API Models}} \\
GPT-5 & 0.8137 & 1.50 & Cloud API \\
Claude-4-Sonnet & 0.7665 & 1.80 & Cloud API \\
\midrule
\multicolumn{4}{l}{\textit{Self-deployment Models}} \\
GPT-OSS-120B & 0.7661 & 1.10 & 32 $\times$ H20 \\
Qwen-2.5-72B & 0.7285 & 1.30 & 32 $\times$ H20 \\
\midrule
\multicolumn{4}{l}{\textit{Our Models}} \\
Qwen-2.5-72B (teacher)$\dagger$ & 0.9497 & 0.186 & 8 $\times$ H20 \\
\textbf{Qwen-3-4B (student)}* & \textbf{0.9599} & \textbf{0.008} & \textbf{8 $\times$ H20} \\
\bottomrule
\end{tabular}
\vspace{1mm}
\end{table}

\subsection{Parameter Analysis}



\begin{figure}[htbp]
    \centering
    \begin{subfigure}{0.48\columnwidth}
        \centering
        \includegraphics[width=\linewidth]{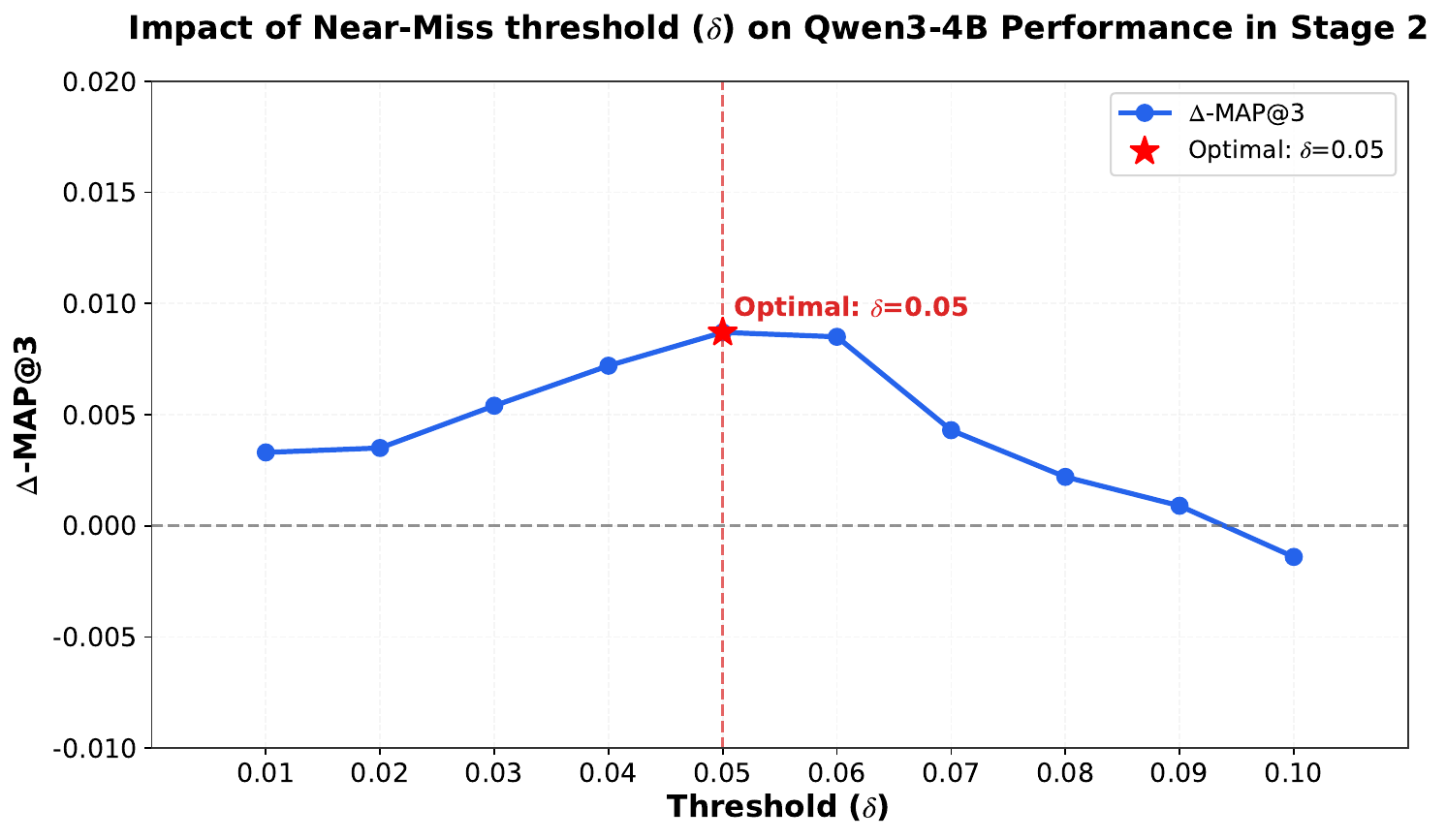}
        \caption{Impact of $\delta$ threshold 
        }
        \label{fig:delta_map3}
    \end{subfigure}
    \hfill
    \begin{subfigure}{0.48\columnwidth}
        \centering
        \includegraphics[width=\linewidth]{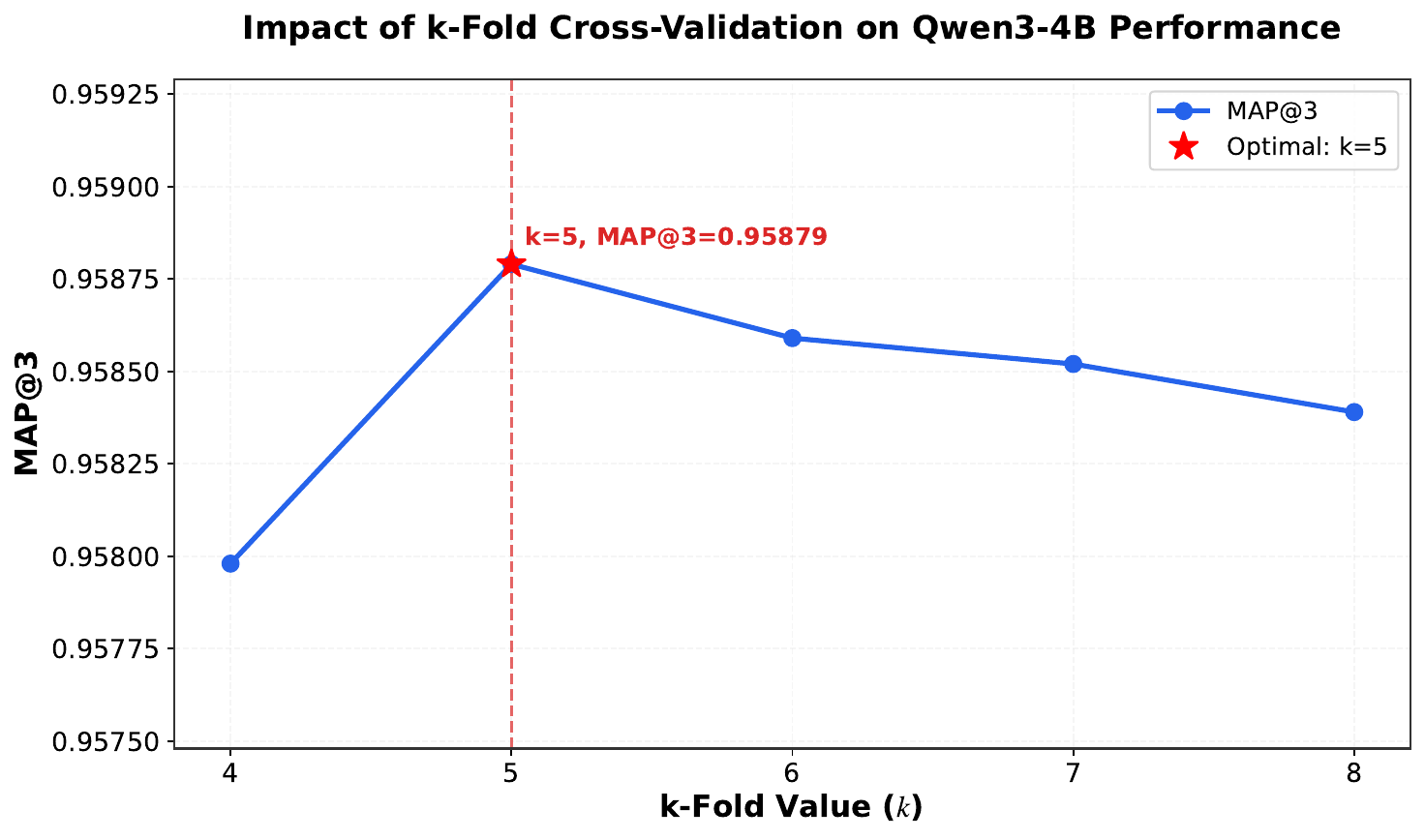}
        \caption{Impact of different K-fold 
        }
        \label{fig:kfold_map3}
    \end{subfigure}

    \caption{Results of two hyperparameter search experiments in the second stage. 
    }
    \label{fig:hyperparam-search}
\end{figure}

In the second training stage, the class probability threshold $\delta$ is used to filter high-uncertainty samples. Based on Qwen-3-4B \cite{qwen3}, experiments in the range $[0.01, 0.10]$ evaluated changes in validation $\mathrm{MAP}@3$, with gain defined as $\mathrm{MAP}@3(\delta) - \mathrm{MAP}@3(\delta_{\mathrm{baseline}})$. Results (Figure~\ref{fig:delta_map3}) show $\delta = 0.05$ yields the largest improvement ($+0.012$), while $\delta = 0.10$ causes a drop ($-0.004$), indicating that moderate thresholds effectively select useful samples, whereas overly high thresholds risk overfitting. K-fold validation further shows performance peaks at $K=5$, achieving \textbf{0.95879} $\mathrm{MAP}@3$, suggesting this split optimally balances training coverage and validation stability.

\section{Conclusion}

This study proposes a hierarchical-difficulty-based sample selection and probability-constrained multi-scale knowledge distillation framework, achieving the high accuracy required for real-world deployment of educational AI under extremely data-scarce conditions.  
Leveraging a dual-level margin mechanism to precisely identify \emph{Near-miss} (correct predictions with uncertainty) and \emph{Hard-hard} (severely incorrect predictions) samples, and designing adaptive loss functions for different types of samples, we achieve a MAP@3 of $0.9585$ (relative improvement of $17.8\%$) on real student data, significantly enhancing the reliability of erroneous concept diagnosis.  
This framework can be extended in the future to progressive difficulty scheduling and multi-task learning, improving the personalization and scalability of automated tutoring systems.

\section*{Limitation}

Although this study demonstrates promising results in multi-stage distillation and high-value sample selection, there are still several limitations:



First, the overhead associated with the $K$-fold partition. Although stratified 5-fold cross-validation helps mitigate overfitting (see Appendix \ref{app:limitations_scale}), we adopt a $K$-fold cross-partition approach to ensure reliability and prevent data leakage. However, searching for the optimal $K$ is complex, as each evaluation requires a complete global training cycle. This high overhead may limit the exploration of optimal configurations in resource-constrained scenarios, thereby constraining the method's deployment potential.

Second, the limited improvement for low-quality data. While our multi-stage distillation framework enhances performance by selecting valuable samples, its efficacy is limited when incoming data is inherently of poor quality. In the presence of large quality gaps, the model cannot achieve significant gains solely through sample selection. Therefore, it is crucial to design effective high-quality data synthesis strategies to actively generate and repair data, rather than relying only on filtering existing samples.

\section*{Acknowledgments}
This work was supported by the ``Pioneer'' and ``Leading Goose'' R\&D Program of Zhejiang (No.~2026C02A1236), the National Natural Science Foundation of China (No.~62577050), and the Jinhua Major Science and Technology Project (No.~2024-1-005).

\bibliography{ref}

@misc{benchmark1,
    author = {Jules King and Kennedy Smith and L Burleigh and Scott Crossley and Maggie Demkin and Walter Reade},
    title = {MAP - Charting Student Math Misunderstandings},
    year = {2025},
    howpublished = {\url{https://kaggle.com/competitions/map-charting-student-math-misunderstandings}},
    note = {Kaggle}
}

@misc{benchmark2,
      title={A Benchmark for Math Misconceptions: Bridging Gaps in Middle School Algebra with AI-Supported Instruction}, 
      author={Otero Nancy and Druga Stefania and Lan Andrew},
      year={2024},
      eprint={2412.03765},
      archivePrefix={arXiv},
      primaryClass={cs.HC},
      url={https://arxiv.org/abs/2412.03765}, 
}

@misc{qwen3,
      title={Qwen3 Technical Report}, 
      author={Qwen Team},
      year={2025},
      eprint={2505.09388},
      archivePrefix={arXiv},
      primaryClass={cs.CL},
      url={https://arxiv.org/abs/2505.09388}, 
}

@misc{qwen2.5,
    title = {Qwen2.5: A Party of Foundation Models},
    url = {https://qwenlm.github.io/blog/qwen2.5/},
    author = {Qwen Team},
    month = {September},
    year = {2024}
}

@article{chanllenge-1,
  author    = {Dyer, Elizabeth B. and Sherin, Miriam Gamoran},
  title     = {Instructional reasoning about interpretations of student thinking that supports responsive teaching in secondary mathematics},
  journal   = {ZDM Mathematics Education},
  year      = {2016},
  volume    = {48},
  number    = {1-2},
  pages     = {69--82},
  doi       = {10.1007/s11858-015-0740-1},
  url       = {https://doi.org/10.1007/s11858-015-0740-1}
}

@article{gemma,
    title={Gemma},
    url={https://www.kaggle.com/m/3301},
    DOI={10.34740/KAGGLE/M/3301},
    publisher={Kaggle},
    author={Gemma Team},
    year={2024}
}

@article{chanllenge-2,
  author    = {Parwati, Ni and Suharta, I.},
  title     = {Effectiveness of the implementation of cognitive conflict strategy assisted by e-service learning to reduce students’ mathematical misconceptions},
  journal   = {International Journal of Emerging Technologies in Learning (iJET)},
  year      = {2020},
  volume    = {15},
  number    = {11},
  pages     = {102--118},
  doi       = {10.3991/ijet.v15i11.12843},
  url       = {https://doi.org/10.3991/ijet.v15i11.12843}
}

@article{misconception-1,
  author    = {Sadler, Philip M. and Sonnert, Gerhard and Coyle, Heather P. and Cook-Smith, Nancy and Miller, Jaimie L.},
  title     = {The influence of teachers’ knowledge on student learning in middle school physical science classrooms},
  journal   = {American Educational Research Journal},
  year      = {2013},
  volume    = {50},
  number    = {5},
  pages     = {1020--1049},
  doi       = {10.3102/0002831213477680},
  url       = {https://doi.org/10.3102/0002831213477680}
}

@misc{llama-3.1,
      title={The Llama 3 Herd of Models}, 
      author={Aaron Grattafiori and Abhimanyu Dubey and Abhinav Jauhri and Abhinav Pandey},
      year={2024},
      eprint={2407.21783},
      archivePrefix={arXiv},
      primaryClass={cs.AI},
      url={https://arxiv.org/abs/2407.21783}, 
}

@misc{k-fold,
      title={Model Evaluation, Model Selection, and Algorithm Selection in Machine Learning}, 
      author={Sebastian Raschka},
      year={2020},
      eprint={1811.12808},
      archivePrefix={arXiv},
      primaryClass={cs.LG},
      url={https://arxiv.org/abs/1811.12808}, 
}

@misc{gpt-5,
    title = {GPT-5 System Card},
    url = {https://openai.com/zh-Hans-CN/index/introducing-gpt-5/},
    author = {OPENAI Team},
    month = {August},
    year = {2025}
}

@misc{claude,
    title = {Introducing Claude 4},
    url = {https://www.anthropic.com/news/claude-4},
    author = {Claude Team},
    month = {May},
    year = {2025}
}

@misc{DeepSeek-V3,
      title={DeepSeek-V3 Technical Report}, 
      author={DeepSeek-AI and Aixin Liu and Bei Feng and Bing Xue and Bingxuan Wang and Bochao Wu and Chengda Lu and Chenggang Zhao and Chengqi Deng and Chenyu Zhang and Chong Ruan and Damai Dai and Daya Guo and Dejian Yang and Deli Chen and Dongjie Ji and Erhang Li and Fangyun Lin and Fucong Dai and Fuli Luo and Guangbo Hao and Guanting Chen and Guowei Li},
      year={2025},
      eprint={2412.19437},
      archivePrefix={arXiv},
      primaryClass={cs.CL},
      url={https://arxiv.org/abs/2412.19437}, 
}

@misc{HIT,
    author = {Carly D. Robinson, Susanna Loeb},
    title = {High-Impact Tutoring: State of the Research and Priorities for Future Learning},
    month = {May},
    year = {2021},
    url={https://edworkingpapers.com/ai21-384}, 
}

@article{p-VAE,
  author       = {Zichao Wang and
                  Sebastian Tschiatschek and
                  Simon Woodhead and
                  Jos{\'{e}} Miguel Hern{\'{a}}ndez{-}Lobato and
                  Simon Peyton Jones and
                  Cheng Zhang},
  title        = {Large-Scale Educational Question Analysis with Partial Variational
                  Auto-encoders},
  journal      = {CoRR},
  volume       = {abs/2003.05980},
  year         = {2020},
  url          = {https://arxiv.org/abs/2003.05980},
  eprinttype    = {arXiv},
  eprint       = {2003.05980},
  bibsource    = {dblp computer science bibliography, https://dblp.org}
}

@article{AutoTutor,
  author    = {Graesser, Arthur C. and Lu, Shulan and Jackson, George Tanner and Mitchell, Heather Hite and Ventura, Mathew and Olney, Andrew and Louwerse, Max M.},
  title     = {AutoTutor: A tutor with dialogue in natural language},
  journal   = {Behavior Research Methods, Instruments, \& Computers},
  year      = {2004},
  month     = {May},
  volume    = {36},
  number    = {2},
  pages     = {180--192},
  doi       = {10.3758/BF03195563},
  url       = {https://doi.org/10.3758/BF03195563},
  issn      = {1532-5970}
}

@article{gsm-ic,
  title={Large Language Models Can Be Easily Distracted by Irrelevant Context},
  author={Shi, Freda and Chen, Xinyun and Misra, Kanishka and Scales, Nathan and Dohan, David and Chi, Ed and Schärli, Nathanael and Zhou, Denny},
  journal={arXiv preprint arXiv:2302.00093},
  year={2023}
}

@misc{data-survey,
      title={Large Language Models for Data Annotation and Synthesis: A Survey}, 
      author={Zhen Tan and Dawei Li and Song Wang and Alimohammad Beigi and Bohan Jiang and Amrita Bhattacharjee and Mansooreh Karami and Jundong Li and Lu Cheng and Huan Liu},
      year={2024},
      eprint={2402.13446},
      archivePrefix={arXiv},
      primaryClass={cs.CL},
      url={https://arxiv.org/abs/2402.13446}, 
}

@InProceedings{MaE,
author="Ansari, Shahina Mohd Azam
and Bywater, James
and Lilly, Sarah
and Brown, Donald
and Chiu, Jennifer",
editor="Cristea, Alexandra I.
and Walker, Erin
and Lu, Yu
and Santos, Olga C.
and Isotani, Seiji",
title="MisstepMath: A Diverse Student Mistake Dataset for AI in Mathematics Teacher Training",
booktitle="Artificial Intelligence in Education",
year="2025",
publisher="Springer Nature Switzerland",
address="Cham",
pages="381--394",
}

@misc{MathEDU,
      title={MathEDU: Towards Adaptive Feedback for Student Mathematical Problem-Solving}, 
      author={Wei-Ling Hsu and Yu-Chien Tang and An-Zi Yen},
      year={2025},
      eprint={2505.18056},
      archivePrefix={arXiv},
      primaryClass={cs.CL},
      url={https://arxiv.org/abs/2505.18056}, 
}

@article{Bridg-gap,
  author    = {Otero, Nancy and Druga, Stefania and Lan, Andrew},
  title     = {A Benchmark for math misconceptions: bridging gaps in middle school algebra with AI-supported instruction},
  journal   = {Discover Education},
  year      = {2025},
  month     = {Aug},
  volume    = {4},
  number    = {1},
  pages     = {277},
  doi       = {10.1007/s44217-025-00742-w},
  url       = {https://doi.org/10.1007/s44217-025-00742-w},
  issn      = {2731-5525},
}

@inproceedings{curriculum,
author = {Bengio, Yoshua and Louradour, J\'{e}r\^{o}me and Collobert, Ronan and Weston, Jason},
title = {Curriculum learning},
year = {2009},
isbn = {9781605585161},
publisher = {Association for Computing Machinery},
address = {New York, NY, USA},
url = {https://doi.org/10.1145/1553374.1553380},
doi = {10.1145/1553374.1553380},
booktitle = {Proceedings of the 26th Annual International Conference on Machine Learning},
pages = {41–48},
numpages = {8},
location = {Montreal, Quebec, Canada},
series = {ICML '09}
}

@misc{self-evolving-curriculumllmreasoning,
      title={Self-Evolving Curriculum for LLM Reasoning}, 
      author={Xiaoyin Chen and Jiarui Lu and Minsu Kim and Dinghuai Zhang and Jian Tang and Alexandre Piché and Nicolas Gontier and Yoshua Bengio and Ehsan Kamalloo},
      year={2025},
      eprint={2505.14970},
      archivePrefix={arXiv},
      primaryClass={cs.AI},
      url={https://arxiv.org/abs/2505.14970}, 
}

@misc{activelearning-1,
      title={Bayesian Active Learning for Classification and Preference Learning}, 
      author={Neil Houlsby and Ferenc Huszár and Zoubin Ghahramani and Máté Lengyel},
      year={2011},
      eprint={1112.5745},
      archivePrefix={arXiv},
      primaryClass={stat.ML},
      url={https://arxiv.org/abs/1112.5745}, 
}

@misc{activelearning-2,
      title={Deep Bayesian Active Learning with Image Data}, 
      author={Yarin Gal and Riashat Islam and Zoubin Ghahramani},
      year={2017},
      eprint={1703.02910},
      archivePrefix={arXiv},
      primaryClass={cs.LG},
      url={https://arxiv.org/abs/1703.02910}, 
}

@misc{activelearning-3,
      title={BatchBALD: Efficient and Diverse Batch Acquisition for Deep Bayesian Active Learning}, 
      author={Andreas Kirsch and Joost van Amersfoort and Yarin Gal},
      year={2019},
      eprint={1906.08158},
      archivePrefix={arXiv},
      primaryClass={cs.LG},
      url={https://arxiv.org/abs/1906.08158}, 
}

@misc{distillation,
      title={Distilling the Knowledge in a Neural Network}, 
      author={Geoffrey Hinton and Oriol Vinyals and Jeff Dean},
      year={2015},
      eprint={1503.02531},
      archivePrefix={arXiv},
      primaryClass={stat.ML},
      url={https://arxiv.org/abs/1503.02531}, 
}

@misc{distillation-survey,
      title={A Comprehensive Survey on Knowledge Distillation}, 
      author={Amir M. Mansourian and Rozhan Ahmadi and Masoud Ghafouri and Amir Mohammad Babaei and Elaheh Badali Golezani and Zeynab Yasamani Ghamchi and Vida Ramezanian and Alireza Taherian and Kimia Dinashi and Amirali Miri and Shohreh Kasaei},
      year={2025},
      eprint={2503.12067},
      archivePrefix={arXiv},
      primaryClass={cs.CV},
      url={https://arxiv.org/abs/2503.12067}, 
}

@misc{adaptive-distillation,
      title={Adaptive Distillation: Aggregating Knowledge from Multiple Paths for Efficient Distillation}, 
      author={Sumanth Chennupati and Mohammad Mahdi Kamani and Zhongwei Cheng and Lin Chen},
      year={2021},
      eprint={2110.09674},
      archivePrefix={arXiv},
      primaryClass={cs.CV},
      url={https://arxiv.org/abs/2110.09674}, 
}

@article{spot-adaptive-distillation,
   title={Spot-Adaptive Knowledge Distillation},
   volume={31},
   ISSN={1941-0042},
   url={http://dx.doi.org/10.1109/TIP.2022.3170728},
   DOI={10.1109/tip.2022.3170728},
   journal={IEEE Transactions on Image Processing},
   publisher={Institute of Electrical and Electronics Engineers (IEEE)},
   author={Song, Jie and Chen, Ying and Ye, Jingwen and Song, Mingli},
   year={2022},
   pages={3359–3370} }

@misc{hard-adaptive-distillation,
      title={Rethinking the Generation of High-Quality CoT Data from the Perspective of LLM-Adaptive Question Difficulty Grading}, 
      author={Qianjin Yu and Keyu Wu and Zihan Chen and Chushu Zhang and Manlin Mei and Lingjun Huang and Fang Tan and Yongsheng Du and Kunlin Liu and Yurui Zhu},
      year={2025},
      eprint={2504.11919},
      archivePrefix={arXiv},
      primaryClass={cs.AI},
      url={https://arxiv.org/abs/2504.11919}, 
}

@misc{distill-sup-0,
      title={One-for-All: Bridge the Gap Between Heterogeneous Architectures in Knowledge Distillation}, 
      author={Zhiwei Hao and Jianyuan Guo and Kai Han and Yehui Tang and Han Hu and Yunhe Wang and Chang Xu},
      year={2023},
      eprint={2310.19444},
      archivePrefix={arXiv},
      primaryClass={cs.CV},
      url={https://arxiv.org/abs/2310.19444}, 
}

@misc{distill-sup-1,
      title={Dataset Distillation by Matching Training Trajectories}, 
      author={George Cazenavette and Tongzhou Wang and Antonio Torralba and Alexei A. Efros and Jun-Yan Zhu},
      year={2022},
      eprint={2203.11932},
      archivePrefix={arXiv},
      primaryClass={cs.CV},
      url={https://arxiv.org/abs/2203.11932}, 
}

@misc{distill-sup-2,
      title={Dataset Distillation via Factorization}, 
      author={Songhua Liu and Kai Wang and Xingyi Yang and Jingwen Ye and Xinchao Wang},
      year={2022},
      eprint={2210.16774},
      archivePrefix={arXiv},
      primaryClass={cs.CV},
      url={https://arxiv.org/abs/2210.16774}, 
}

@misc{dis-sup-3,
      title={MSD: Multi-Self-Distillation Learning via Multi-classifiers within Deep Neural Networks}, 
      author={Yunteng Luan and Hanyu Zhao and Zhi Yang and Yafei Dai},
      year={2019},
      eprint={1911.09418},
      archivePrefix={arXiv},
      primaryClass={cs.CV},
      url={https://arxiv.org/abs/1911.09418}, 
}

@misc{dis-sup-4,
      title={LumiNet: Perception-Driven Knowledge Distillation via Statistical Logit Calibration}, 
      author={Md. Ismail Hossain and M M Lutfe Elahi and Sameera Ramasinghe and Ali Cheraghian and Fuad Rahman and Nabeel Mohammed and Shafin Rahman},
      year={2025},
      eprint={2310.03669},
      archivePrefix={arXiv},
      primaryClass={cs.CV},
      url={https://arxiv.org/abs/2310.03669}, 
}

@misc{dis-sup-5,
      title={Knowledge Distillation with the Reused Teacher Classifier}, 
      author={Defang Chen and Jian-Ping Mei and Hailin Zhang and Can Wang and Yan Feng and Chun Chen},
      year={2022},
      eprint={2203.14001},
      archivePrefix={arXiv},
      primaryClass={cs.CV},
      url={https://arxiv.org/abs/2203.14001}, 
}

@InProceedings{dis-sup-6,
author = {Guo, Qiushan and Wang, Xinjiang and Wu, Yichao and Yu, Zhipeng and Liang, Ding and Hu, Xiaolin and Luo, Ping},
title = {Online Knowledge Distillation via Collaborative Learning},
booktitle = {Proceedings of the IEEE/CVF Conference on Computer Vision and Pattern Recognition (CVPR)},
month = {June},
year = {2020}
}

@misc{dis-aug-1,
      title={Reinforced Multi-Teacher Selection for Knowledge Distillation}, 
      author={Fei Yuan and Linjun Shou and Jian Pei and Wutao Lin and Ming Gong and Yan Fu and Daxin Jiang},
      year={2020},
      eprint={2012.06048},
      archivePrefix={arXiv},
      primaryClass={cs.CL},
      url={https://arxiv.org/abs/2012.06048}, 
}

@misc{dis-aug-2,
      title={Contrastive Model Inversion for Data-Free Knowledge Distillation}, 
      author={Gongfan Fang and Jie Song and Xinchao Wang and Chengchao Shen and Xingen Wang and Mingli Song},
      year={2021},
      eprint={2105.08584},
      archivePrefix={arXiv},
      primaryClass={cs.AI},
      url={https://arxiv.org/abs/2105.08584}, 
}

@misc{dis-aug-3,
      title={Dreaming to Distill: Data-free Knowledge Transfer via DeepInversion}, 
      author={Hongxu Yin and Pavlo Molchanov and Zhizhong Li and Jose M. Alvarez and Arun Mallya and Derek Hoiem and Niraj K. Jha and Jan Kautz},
      year={2020},
      eprint={1912.08795},
      archivePrefix={arXiv},
      primaryClass={cs.LG},
      url={https://arxiv.org/abs/1912.08795}, 
}

\clearpage
\appendix

\section{Adaptive Loss Weight Allocation}
\label{sec:loss_weights}

Table~\ref{tab:loss_weights} presents the specific weighting strategy for the adaptive loss function based on sample categories. This mechanism organically combines sample difficulty characterization with label credibility assessment, and in particular, demonstrates a careful handling of real-world noise for $\mathcal{S}_{\mathrm{HH}}^{\mathrm{close}}$ samples. This is a key design choice that enhances model robustness in complex categories.

\begin{table}[H]
\centering
\caption{Adaptive loss weight allocation strategy based on sample categories}
\label{tab:loss_weights}
\begin{tabular}{@{}lccc@{}}
\toprule
Category & $\alpha (\mathcal{L}_{\text{CE}})$ & $\beta (\mathcal{L}_{\text{KD}})$ & $\gamma (\mathcal{L}_{\text{COS}})$ \\
\midrule
$\mathcal{S}_{\text{NM}}^{\text{close}}$ & 1.0 & 0.0 & 0.0  \\
$\mathcal{S}_{\text{NM}}^{\text{far}}$ & 1.0 & 1.0 & 1.0  \\
$\mathcal{S}_{\text{HH}}^{\text{close}}$ & 0.0 & 1.0 & 1.0  \\
$\mathcal{S}_{\text{HH}}^{\text{far}}$ & 1.0 & 1.0 & 1.0 \\
\bottomrule
\end{tabular}
\end{table}

\section{Detailed Error Analysis Across Categories}
\label{sec:error-analysis-appendix}

\begin{figure}[htbp]
    \centering
    \begin{subfigure}{0.48\columnwidth}
        \centering
        \includegraphics[width=\linewidth]{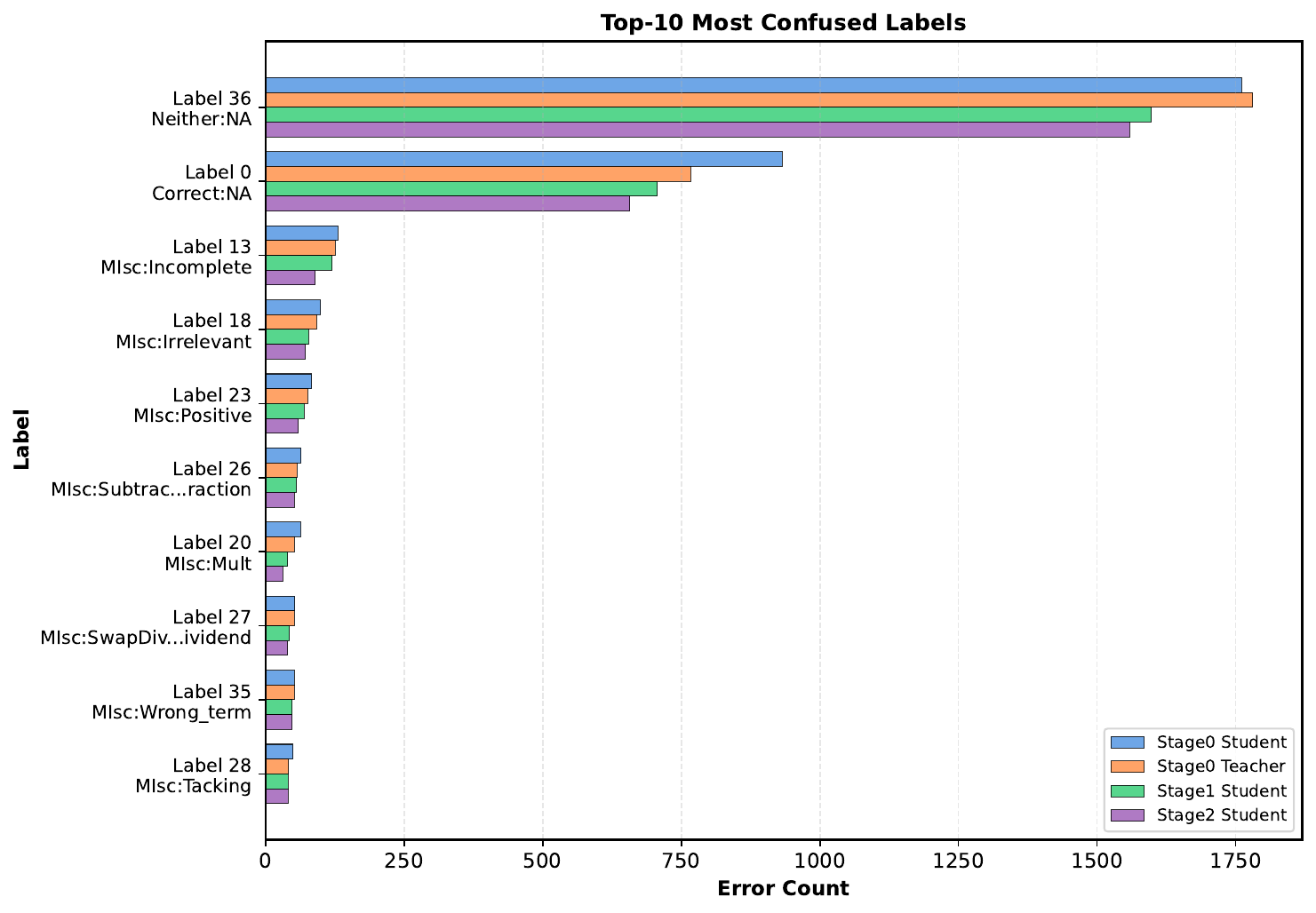}
        \caption{Error count changes in top-10 confused categories}
        \label{fig:error-top10}
    \end{subfigure}
    \hfill
    \begin{subfigure}{0.48\columnwidth}
        \centering
        \includegraphics[width=\linewidth]{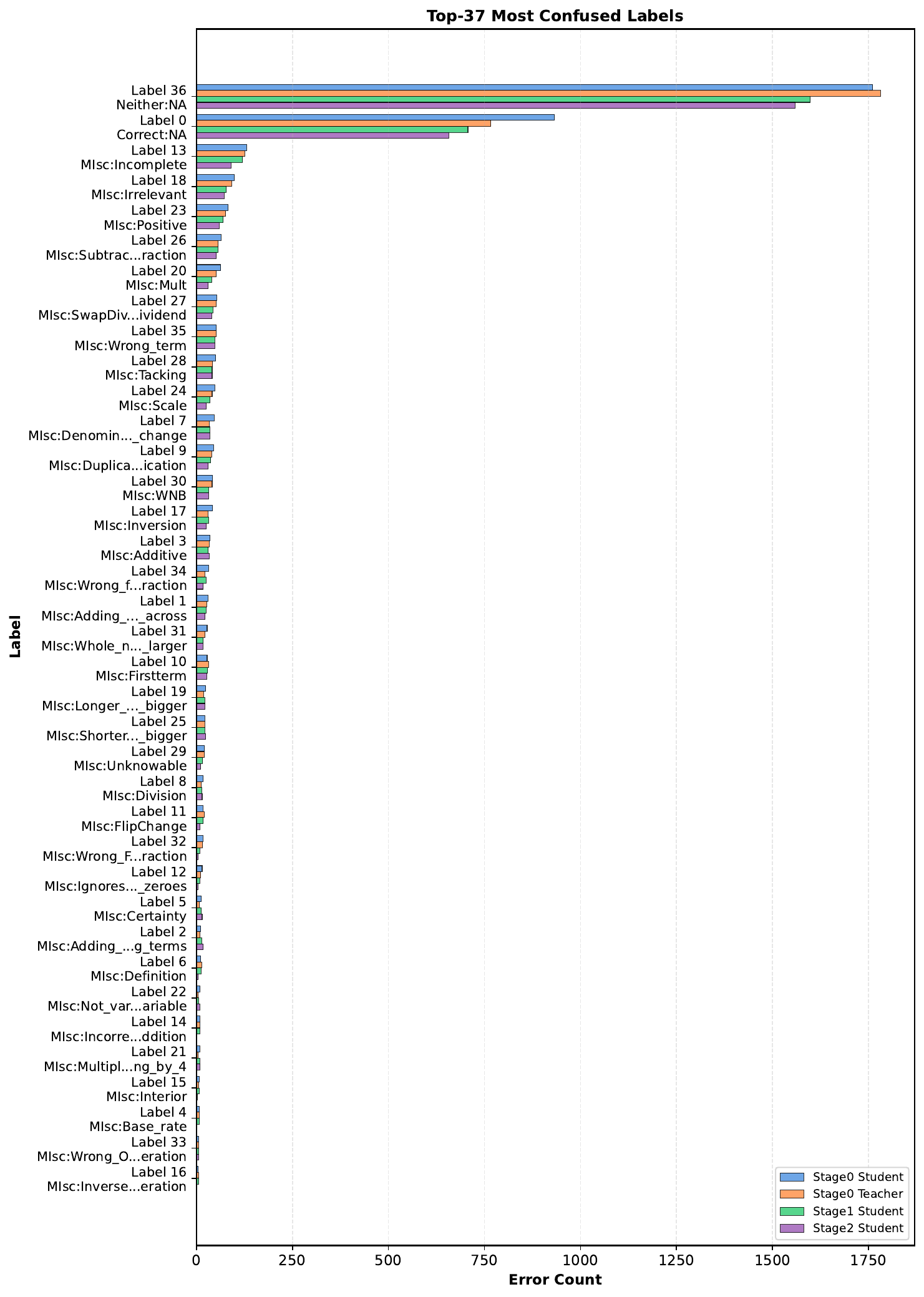}
        \caption{Error count changes in all 37 categories}
        \label{fig:error-all}
    \end{subfigure}

    \caption{Extended error analysis for different category groups}
    \label{fig:error-analysis-complete}
\end{figure}

This section provides comprehensive error analysis results for different numbers of categories. Figure~\ref{fig:error-analysis-complete} shows the error count changes for the top-10 most commonly confused categories and all 37 categories across different stage models.

The analysis reveals that the multi-stage distillation approach consistently reduces errors across all category groups. The improvement is particularly notable in the most frequently misclassified categories, where later-stage models demonstrate substantial error reduction compared to earlier stages.

\section{LLM Model Configuration Details}
\label{sec:llm-configs}

Tables~\ref{tab:llm-sampling-params-1},~\ref{tab:llm-sampling-params-2} and~\ref{tab:llm-inference-params} present the sampling parameters and inference configuration details for all state-of-the-art language models evaluated in our benchmark experiments.

\begin{table}[H]
\centering
\caption{Sampling parameters for state-of-the-art language models (Part 1)}
\label{tab:llm-sampling-params-1}
\resizebox{\columnwidth}{!}{
\begin{tabular}{@{}lccc@{}}
\toprule
\textbf{Model} & \textbf{Temperature} & \textbf{Max Tokens} & \textbf{Top-p} \\
\midrule
GPT-5 & 0.7 & 8192 & 0.9 \\
DeepSeek-V3 & 0.3 & 8192 & 0.9 \\
GPT-OSS-120B & 0.7 & 8192 & 0.9 \\
Claude Sonnet 4 & 0.3 & 8192 & 0.9 \\
Qwen2.5-72B & 0.7 & 8192 & 0.9 \\
\bottomrule
\end{tabular}
}
\end{table}

\begin{table}[H]
\centering
\caption{Sampling parameters for state-of-the-art language models (Part 2)}
\label{tab:llm-sampling-params-2}
\resizebox{\columnwidth}{!}{
\begin{tabular}{@{}lccc@{}}
\toprule
\textbf{Model} & \textbf{Repetition Penalty} & \textbf{Frequency Penalty} & \textbf{Presence Penalty} \\
\midrule
GPT-5 & -- & 0 & 0 \\
DeepSeek-V3 & 1.05 & 0 & 0 \\
GPT-OSS-120B & -- & 0 & 0 \\
Claude Sonnet 4 & 1.05 & 0 & 0 \\
Qwen2.5-72B & 1.05 & 0 & 0 \\
\bottomrule
\end{tabular}
}
\end{table}

\begin{table}[H]
\centering
\caption{Inference configuration for state-of-the-art language models}
\label{tab:llm-inference-params}
\resizebox{\columnwidth}{!}{
\begin{tabular}{@{}lcc@{}}
\toprule
\textbf{Model} & \textbf{Batch Size} & \textbf{Rate Limit} \\
\midrule
GPT-5 & 500 & 2000 \\
DeepSeek-V3 & 25 & 500 \\
GPT-OSS-120B & 500 & 5000 \\
Claude Sonnet 4 & 500 & 2000 \\
Qwen2.5-72B & 50 & 1000 \\
\bottomrule
\end{tabular}
}
\end{table}



\section{Hyperparameter-Setting and Cross-Setting Generalization}
\label{app:hparam_generalization}

\paragraph{Stage-1 Loss Weights.}
To facilitate reproducibility, we report our Stage-1 loss as a weighted combination of cross-entropy (CE), knowledge distillation (KD), and cosine similarity (COS):
$\mathcal{L}=\alpha \mathcal{L}_{\mathrm{CE}} + \beta \mathcal{L}_{\mathrm{KD}} + \gamma \mathcal{L}_{\mathrm{COS}}$, where $\alpha+\beta+\gamma=1$.
We performed a grid-style ablation across three student backbones. A balanced configuration
$(\alpha,\beta,\gamma)=(0.33,0.33,0.34)$ (approximately $1{:}1{:}1$) consistently yields the best (or near-best) performance across architectures and scales, suggesting that our hyperparameter choice is stable rather than overfitted.

\begin{table}[H]
\centering
\scriptsize
\setlength{\tabcolsep}{8pt}
\begin{tabular}{ccccc}
\toprule
$\alpha$ (CE) & $\beta$ (KD) & $\gamma$ (COS) & MAP@3 & Acc. \\
\midrule
0.25 & 0.25 & 0.50 & 0.9476 & 0.8988 \\
0.25 & 0.50 & 0.25 & 0.9466 & 0.8969 \\
0.50 & 0.25 & 0.25 & 0.9445 & 0.8927 \\
0.20 & 0.20 & 0.60 & 0.9476 & 0.8991 \\
0.20 & 0.60 & 0.20 & 0.9489 & 0.9013 \\
0.60 & 0.20 & 0.20 & 0.9414 & 0.8870 \\
0.20 & 0.40 & 0.40 & 0.9493 & 0.9020 \\
0.40 & 0.20 & 0.40 & 0.9452 & 0.8943 \\
0.40 & 0.40 & 0.20 & 0.9459 & 0.8958 \\
\textbf{0.33} & \textbf{0.33} & \textbf{0.34} & \textbf{0.9495} & \textbf{0.9024} \\
0.00 & 0.50 & 0.50 & 0.9478 & 0.8995 \\
0.50 & 0.00 & 0.50 & 0.9420 & 0.8886 \\
0.50 & 0.50 & 0.00 & 0.9466 & 0.8969 \\
0.00 & 0.00 & 1.00 & 0.9360 & 0.8838 \\
0.00 & 1.00 & 0.00 & 0.9488 & 0.9010 \\
1.00 & 0.00 & 0.00 & 0.9386 & 0.8821 \\
\bottomrule
\end{tabular}
\caption{Stage-1 loss weight ablation on Qwen-3-4B.}
\label{tab:stage1_qwen_ablation}
\end{table}

\begin{table}[H]
\centering
\scriptsize
\setlength{\tabcolsep}{8pt}
\begin{tabular}{ccccc}
\toprule
$\alpha$ (CE) & $\beta$ (KD) & $\gamma$ (COS) & MAP@3 & Acc. \\
\midrule
0.25 & 0.25 & 0.50 & 0.9452 & 0.8940 \\
0.25 & 0.50 & 0.25 & 0.9453 & 0.8942 \\
0.50 & 0.25 & 0.25 & 0.9434 & 0.8903 \\
0.20 & 0.20 & 0.60 & 0.9451 & 0.8938 \\
0.20 & 0.60 & 0.20 & 0.9460 & 0.8957 \\
0.60 & 0.20 & 0.20 & 0.9402 & 0.8846 \\
0.20 & 0.40 & 0.40 & 0.9470 & 0.8975 \\
0.40 & 0.20 & 0.40 & 0.9417 & 0.8873 \\
0.40 & 0.40 & 0.20 & 0.9450 & 0.8931 \\
\textbf{0.33} & \textbf{0.33} & \textbf{0.34} & \textbf{0.9474} & \textbf{0.8981} \\
0.00 & 0.50 & 0.50 & 0.9461 & 0.8961 \\
0.50 & 0.00 & 0.50 & 0.9394 & 0.8836 \\
0.50 & 0.50 & 0.00 & 0.9444 & 0.8924 \\
0.00 & 0.00 & 1.00 & 0.5130 & 0.2904 \\
0.00 & 1.00 & 0.00 & 0.9468 & 0.8969 \\
1.00 & 0.00 & 0.00 & 0.9353 & 0.8753 \\
\bottomrule
\end{tabular}
\caption{Stage-1 loss weight ablation on Gemma-2-9B.}
\label{tab:stage1_gemma_ablation}
\end{table}

\begin{table}[H]
\centering
\scriptsize
\setlength{\tabcolsep}{8pt}
\begin{tabular}{ccccc}
\toprule
$\alpha$ (CE) & $\beta$ (KD) & $\gamma$ (COS) & MAP@3 & Acc. \\
\midrule
0.25 & 0.25 & 0.50 & 0.9447 & 0.8928 \\
0.25 & 0.50 & 0.25 & 0.9464 & 0.8962 \\
0.50 & 0.25 & 0.25 & 0.9419 & 0.8876 \\
0.20 & 0.20 & 0.60 & 0.9446 & 0.8934 \\
0.20 & 0.60 & 0.20 & 0.9463 & 0.8958 \\
0.60 & 0.20 & 0.20 & 0.9414 & 0.8868 \\
0.20 & 0.40 & 0.40 & 0.9458 & 0.8955 \\
0.40 & 0.20 & 0.40 & 0.9425 & 0.8891 \\
0.40 & 0.40 & 0.20 & 0.9433 & 0.8901 \\
\textbf{0.33} & \textbf{0.33} & \textbf{0.34} & \textbf{0.9467} & \textbf{0.8971} \\
0.00 & 0.50 & 0.50 & 0.9457 & 0.8951 \\
0.50 & 0.00 & 0.50 & 0.9399 & 0.8846 \\
0.50 & 0.50 & 0.00 & 0.9431 & 0.8901 \\
0.00 & 0.00 & 1.00 & 0.4201 & 0.3957 \\
0.00 & 1.00 & 0.00 & 0.9466 & 0.8968 \\
1.00 & 0.00 & 0.00 & 0.9367 & 0.8788 \\
\bottomrule
\end{tabular}
\caption{Stage-1 loss weight ablation on Llama-3.1-8B.}
\label{tab:stage1_llama_ablation}
\end{table}

\paragraph{Stage-2 Adaptive Distillation.}
\label{app:stage2_adaptive}

In Stage-2, we categorize high-value samples (e.g., NM vs.\ HH) and apply targeted loss compositions.
Across three backbones, the complete adaptive strategy consistently outperforms uniform loss designs.

\begin{table}[H]
\centering
\scriptsize
\setlength{\tabcolsep}{12pt}
\begin{tabular}{lcc}
\toprule
Method (Qwen-3-4B) & MAP@3 & Acc. \\
\midrule
All selected: CE only & 0.9521 & 0.9085 \\
All selected: CE+KD+COS & 0.9536 & 0.9117 \\
NM: CE; HH: KD+COS & 0.9540 & 0.9123 \\
NM: CE+KD+COS; HH: KD+COS & 0.9574 & 0.9178 \\
\textbf{Complete method} & \textbf{0.9585} & \textbf{0.9198} \\
\bottomrule
\end{tabular}
\caption{Stage-2 ablation on Qwen-3-4B.}
\label{tab:stage2_qwen_ablation}
\end{table}

\begin{table}[H]
\centering
\scriptsize
\setlength{\tabcolsep}{12pt}
\begin{tabular}{lcc}
\toprule
Method (Gemma-2-9B) & MAP@3 & Acc. \\
\midrule
All selected: CE only & 0.9503 & 0.9024 \\
All selected: CE+KD+COS & 0.9516 & 0.9107 \\
NM: CE; HH: KD+COS & 0.9524 & 0.9167 \\
NM: CE+KD+COS; HH: KD+COS & 0.9550 & 0.9139 \\
\textbf{Complete method} & \textbf{0.9560} & \textbf{0.9148} \\
\bottomrule
\end{tabular}
\caption{Stage-2 ablation on Gemma-2-9B.}
\label{tab:stage2_gemma_ablation}
\end{table}

\begin{table}[H]
\centering
\scriptsize
\setlength{\tabcolsep}{12pt}
\begin{tabular}{lcc}
\toprule
Method (Llama-3.1-8B) & MAP@3 & Acc. \\
\midrule
All selected: CE only & 0.9502 & 0.9003 \\
All selected: CE+KD+COS & 0.9514 & 0.9067 \\
NM: CE; HH: KD+COS & 0.9514 & 0.9109 \\
NM: CE+KD+COS; HH: KD+COS & 0.9548 & 0.9124 \\
\textbf{Complete method} & \textbf{0.9553} & \textbf{0.9134} \\
\bottomrule
\end{tabular}
\caption{Stage-2 ablation on Llama-3.1-8B.}
\label{tab:stage2_llama_ablation}
\end{table}

Our adaptive approach that categorizes high-value samples into four types and applies targeted loss combinations consistently outperforms uniform loss strategies across all models, further demonstrating the robustness of our method.
These extensive ablation studies demonstrate that: (1) our Stage 1 hyperparameters generalize well across different model architectures, and (2) our Stage 2 adaptive strategy provides consistent improvements. The high degree of cross-model consistency suggests that our hyperparameter choices are principled rather than overfitted to specific datasets.


\paragraph{Cross-validation for generalization.}
We report full K-fold(K=5) results under the selected balanced setting $(\alpha,\beta,\gamma)=(0.33,0.33,0.34)$.

\begin{table}[H]
\centering
\scriptsize
\setlength{\tabcolsep}{4pt}
\begin{tabular}{lccccc}
\toprule
Fold & $\alpha$ & $\beta$ & $\gamma$ & MAP@3 & Acc. \\
\midrule
fold0 & 0.33 & 0.33 & 0.34 & 0.9495 & 0.9024 \\
fold1 & 0.33 & 0.33 & 0.34 & 0.9499 & 0.9020 \\
fold2 & 0.33 & 0.33 & 0.34 & 0.9463 & 0.8962 \\
fold3 & 0.33 & 0.33 & 0.34 & 0.9464 & 0.8969 \\
fold4 & 0.33 & 0.33 & 0.34 & 0.9490 & 0.9013 \\
\midrule
\textbf{Mean$\pm$Std} & -- & -- & -- & \textbf{0.9482$\pm$0.0017} & \textbf{0.8998$\pm$0.0028} \\
\bottomrule
\end{tabular}
\caption{Complete 5-fold results for Stage-1 (Qwen-3-4B).}
\label{tab:stage1_kfold_qwen_complete}
\end{table}

\begin{table}[H]
\centering
\scriptsize
\setlength{\tabcolsep}{4pt}
\begin{tabular}{lccccc}
\toprule
Fold & $\alpha$ & $\beta$ & $\gamma$ & MAP@3 & Acc. \\
\midrule
fold0 & 0.33 & 0.33 & 0.34 & 0.9474 & 0.8981 \\
fold1 & 0.33 & 0.33 & 0.34 & 0.9480 & 0.8986 \\
fold2 & 0.33 & 0.33 & 0.34 & 0.9457 & 0.8947 \\
fold3 & 0.33 & 0.33 & 0.34 & 0.9465 & 0.8968 \\
fold4 & 0.33 & 0.33 & 0.34 & 0.9474 & 0.8973 \\
\midrule
\textbf{Mean$\pm$Std} & -- & -- & -- & \textbf{0.9470$\pm$0.0009} & \textbf{0.8971$\pm$0.0015} \\
\bottomrule
\end{tabular}
\caption{Complete 5-fold results for Stage-1 (Gemma-2-9B).}
\label{tab:stage1_kfold_gemma_complete}
\end{table}

\begin{table}[H]
\centering
\scriptsize
\setlength{\tabcolsep}{4pt}
\begin{tabular}{lccccc}
\toprule
Fold & $\alpha$ & $\beta$ & $\gamma$ & MAP@3 & Acc. \\
\midrule
fold0 & 0.33 & 0.33 & 0.34 & 0.9467 & 0.8971 \\
fold1 & 0.33 & 0.33 & 0.34 & 0.9455 & 0.8943 \\
fold2 & 0.33 & 0.33 & 0.34 & 0.9445 & 0.8907 \\
fold3 & 0.33 & 0.33 & 0.34 & 0.9462 & 0.8972 \\
fold4 & 0.33 & 0.33 & 0.34 & 0.9465 & 0.8966 \\
\midrule
\textbf{Mean$\pm$Std} & -- & -- & -- & \textbf{0.9459$\pm$0.0009} & \textbf{0.8952$\pm$0.0026} \\
\bottomrule
\end{tabular}
\caption{Complete 5-fold results for Stage-1 (Llama-3.1-8B).}
\label{tab:stage1_kfold_llama_complete}
\end{table}


\begin{table}[H]
\centering
\scriptsize
\setlength{\tabcolsep}{5pt}
\begin{tabular}{lcc}
\toprule
Method (Qwen-3-4B) & MAP@3 & Acc. \\
\midrule
fold0\_exp5 (Complete method) & 0.95845 & 0.91982 \\
fold1\_exp5 & 0.95911 & 0.92006 \\
fold2\_exp5 & 0.95692 & 0.91745 \\
fold3\_exp5 & 0.95877 & 0.91849 \\
fold4\_exp5 & 0.96031 & 0.92147 \\
\midrule
\textbf{Mean$\pm$Std} & \textbf{0.9567$\pm$0.0013} & \textbf{0.9195$\pm$0.0016} \\
\bottomrule
\end{tabular}
\caption{Complete 5-fold results for Stage-2 (Qwen-3-4B), complete method (exp5).}
\label{tab:stage2_kfold_qwen_complete}
\end{table}

\begin{table}[H]
\centering
\scriptsize
\setlength{\tabcolsep}{5pt}
\begin{tabular}{lcc}
\toprule
Method (Gemma-2-9B) & MAP@3 & Acc. \\
\midrule
fold0\_exp5 (Complete method) & 0.95600 & 0.91475 \\
fold1\_exp5 & 0.95798 & 0.91681 \\
fold2\_exp5 & 0.95658 & 0.91573 \\
fold3\_exp5 & 0.95731 & 0.91713 \\
fold4\_exp5 & 0.95827 & 0.91893 \\
\midrule
\textbf{Mean$\pm$Std} & \textbf{0.9572$\pm$0.0009} & \textbf{0.9167$\pm$0.0014} \\
\bottomrule
\end{tabular}
\caption{Complete 5-fold results for Stage-2 (Gemma-2-9B), complete method (exp5).}
\label{tab:stage2_kfold_gemma_complete}
\end{table}

\begin{table}[H]
\centering
\scriptsize
\setlength{\tabcolsep}{5pt}
\begin{tabular}{lcc}
\toprule
Method (Llama-3.1-8B) & MAP@3 & Acc. \\
\midrule
fold0\_exp5 (Complete method) & 0.95530 & 0.91338 \\
fold1\_exp5 & 0.95668 & 0.91574 \\
fold2\_exp5 & 0.95570 & 0.91328 \\
fold3\_exp5 & 0.95682 & 0.91403 \\
fold4\_exp5 & 0.95834 & 0.91867 \\
\midrule
\textbf{Mean$\pm$Std} & \textbf{0.9566$\pm$0.0011} & \textbf{0.9150$\pm$0.0021} \\
\bottomrule
\end{tabular}
\caption{Complete 5-fold results for Stage-2 (Llama-3.1-8B), complete method (exp5).}
\label{tab:stage2_kfold_llama_complete}
\end{table}

The consistent high performance across all folds and different model architectures validates the generalizabilit of our approach, while the small-dataset experiments confirm its practicality.

\section{Dependence on Teacher-Model Uncertainty}
\label{app:teacher_uncertainty}

Our framework utilizes teacher cognitive uncertainty as a \emph{guiding signal} rather than a hard constraint. It serves to identify high-value samples and modulate the student's reliance via difficulty-adaptive weighting. Crucially, when teacher signals are unreliable, the adaptive objective reduces their contribution in favor of ground-truth supervision, effectively preventing the inheritance of teacher flaws.

\section{Comparisons/Visualizations of Selected High-Value Samples}
\label{app:sample_visualization}

We provide side-by-side, box-rendered examples to illustrate how our selection distinguishes low-value vs.\ high-value samples and why different sample types require different supervision.


\newtcolorbox{SampleBox}[2][]{%
  enhanced,
  breakable,
  colback=gray!6,
  colframe=black!35,
  boxrule=0.6pt,
  arc=2pt,
  left=6pt,right=6pt,top=4pt,bottom=4pt,
  fonttitle=\bfseries,
  title={#2},
  #1
}

\begin{SampleBox}[colback=green!6,colframe=green!55!black,colbacktitle=green!18,coltitle=black]{Case 1: Easy/Normal Sample (Low Learning Value)}
{\small\ttfamily
\textbf{QuestionText:} What fraction of the shape is not shaded? Give your answer in its simplest form.\\
\textbf{MC\_Answer:} 1/3\\
\textbf{StudentExplanation:} "one third is equal to tree nineth"\\
\textbf{pred\_top3:} [0, 36, 13]\\
\textbf{pred\_probability:} [0.910085, 0.060027, 0.002885]\\
\textbf{label:} 0\\
\noindent\textbf{Interpretation:} The teacher predicts correctly with high confidence (91\%), indicating limited marginal training value.
}
\vspace{2pt}

\end{SampleBox}

\begin{SampleBox}[colback=orange!8,colframe=orange!70!black,colbacktitle=orange!18,coltitle=black]{Case 2: NM-close (Borderline Confusion)}
{\small\ttfamily
\textbf{QuestionText:} What fraction of the shape is not shaded? Give your answer in its simplest form.\\
\textbf{MC\_Answer:} 1/3\\
\textbf{StudentExplanation:} "Because its simplified from 3 ninth"\\
\textbf{pred\_top3:} [36, 0, 13]\\
\textbf{pred\_probability:} [0.532756, 0.428080, 0.004627]\\
\textbf{label:} 0\\
\noindent\textbf{Interpretation:} Probabilities are close (decision boundary). These samples benefit from stronger ground-truth supervision (CE) to sharpen the boundary.
}
\vspace{2pt}

\end{SampleBox}

\begin{SampleBox}[colback=red!6,colframe=red!65!black,colbacktitle=red!16,coltitle=black]{Case 3: NM-far (High-Confidence Misjudgment / Teacher Bias)}
{\small\ttfamily
\textbf{QuestionText:} What fraction of the shape is not shaded? Give your answer in its simplest form.\\
\textbf{MC\_Answer:} 1/3\\
\textbf{StudentExplanation:} "Because there are nine 3rds all together so simply that"\\
\textbf{pred\_top3:} [36, 0, 13]\\
\textbf{pred\_probability:} [0.936680, 0.043809, 0.003186]\\
\textbf{label:} 0\\
\noindent\textbf{Interpretation.} The teacher is confidently wrong. Joint supervision (ground truth + teacher representation) helps the student learn semantic features while correcting the teacher’s mistaken label preference.
}
\vspace{2pt}

\end{SampleBox}

\begin{SampleBox}[colback=blue!6,colframe=blue!60!black,colbacktitle=blue!16,coltitle=black]{Case 4: HH-near (Complex Semantics / Confusable Error Types)}
{\small\ttfamily
\textbf{QuestionText:} Calculate 1/2 $\div$ 6\\
\textbf{MC\_Answer:} 3\\
\textbf{StudentExplanation:} "I think this because half of 6 is 3 and 2 divided by 6 is 3."\\
\textbf{pred\_top3:} [20, 11, 27]\\
\textbf{pred\_probability:} [0.877975, 0.029347, 0.017252]\\
\textbf{label:} 36\\
\noindent\textbf{Interpretation.} The explanation superficially matches multiple misconceptions, making it easy to confuse. For such samples, we prioritize teacher signals (KD+COS) to transfer richer semantic discrimination.
}
\vspace{2pt}

\end{SampleBox}

\begin{SampleBox}[colback=violet!7,colframe=violet!60!black,colbacktitle=violet!18,coltitle=black]{Case 5: HH-far (Diverse Expression / Teacher Blind Spot)}
{\small\ttfamily
\textbf{QuestionText:} A/10 = 9/15. What is the value of A?\\
\textbf{MC\_Answer:} 1/3\\
\textbf{StudentExplanation:} "because half is added so we got 9 from 6"\\
\textbf{pred\_top3:} [36, 3, 10]\\
\textbf{pred\_probability:} [0.912709, 0.037672, 0.008021]\\
\textbf{label:} 0\\
\noindent\textbf{Interpretation.} The teacher misjudges unconventional yet potentially valid reasoning with high confidence. These samples expose teacher blind spots; difficulty-adaptive weighting increases reliance on ground truth, enabling the student to surpass the teacher.
}
\vspace{2pt}

\end{SampleBox}

\section{Why the Student Can Outperform the Teacher}
\label{app:student_beats_teacher}

We observed that a smaller student can match or surpass a larger teacher on this task due to:
\begin{itemize}
  \item \textbf{Teacher pretraining bias and cognitive blind spots.} Large general-purpose teachers may over-prefer standardized reasoning, while student explanations are often non-standard but self-consistent.
  \item \textbf{Task specialization via high-value selection.} Our selection focuses learning on uncertainty-revealing regions that are most diagnostic for misconception classification.
  \item \textbf{Adaptive correction of teacher errors.} When the teacher is confidently wrong (e.g., NM-far / HH-far), difficulty-adaptive weighting increases the contribution of ground-truth supervision, enabling the student to inherit general knowledge while correcting teacher-specific mistakes.
\end{itemize}

\section{Limitations: Small Dataset Scale and Data Availability}
\label{app:limitations_scale}

A key limitation is the \textbf{difficulty of collecting authentic student reasoning at scale}. While we mitigate overfitting concerns through stratified 5-fold validation on the 36k dataset and additionally verify practicality on a smaller curated set (220 samples), broader generalization is still constrained by:
(i) limited public datasets that contain authentic student explanations with sufficient quality, and
(ii) the inherent cost of obtaining large-scale real-world student reasoning.
We view expanding dataset coverage (languages, curricula, demographics) as important future work.


\section{Prompt for Experiments}


To validate the effectiveness of our method, we designed three different experimental scenarios, with corresponding prompts shown below.
Through these three different types of prompts, we can comprehensively evaluate the model’s performance under both data synthesis and direct prediction modes.


\onecolumn
\begin{tcolorbox}[
    colback=blue!5!white,
    colframe=blue!75!black,
    title=\textbf{Prompt: Synthetic Student Explanation Generator},
    fonttitle=\bfseries,
    breakable,
    enhanced,
    segmentation style={draw=none},
    width=\textwidth,
    left=2mm,
    right=2mm,
    top=2mm,
    bottom=2mm,
    label=fig:synthetic-prompt
]
\small\ttfamily
You will be generating new student explanations that match a specific explanation type for a 
given math question. You need to simulate how different students at the same grade level might 
explain their reasoning when arriving at the same answer.

\textbf{Here is the question, student's answer, answer correctness and an example student explanation:}

<question>
\{QUESTIONTEXT\}
</question> \\

<answer>
\{ANSWER\}
</answer> \\

<answer\_correctness>
\{ANSWER CORRECTNESS\}
</answer\_correctness> \\

<student\_explanation>
\{STUDENT EXPLANATION\}
</student\_explanation> \\

\textbf{Here is the explanation type you need to match and related student explanations from other students:}

<explanation\_type>\\
\{STUDENT\_EXPLANATION\_TYPE\}\\
</explanation\_type> \\

<related\_explanations>\\
\{RELATED\_STUDENT\_EXPLANATION\}\\
</related\_explanations> \\

You need to generate \{N\} new student explanations.\\

\textbf{\#\# EXPLANATION TYPE CATEGORIES:}\\

- \textbf{Correct:NA} - The student's reasoning process is mathematically sound and leads 
  logically to the correct answer
  
- \textbf{Neither:NA} - The student's explanation is vague, unclear, or unrelated to the 
  mathematical concepts in the question
  
- \textbf{Misconception:[Specific type]} - The student has a specific mathematical misconception. 
  The specific type describes what mathematical concept they misunderstand 
  (e.g., "Incorrect\_equivalent\_fraction\_addition") \\

\textbf{\#\# INSTRUCTIONS:}\\

1. Analyze the given explanation type to understand what kind of reasoning pattern you need 
   to replicate
   
2. Study the related student explanations to understand the common patterns for this 
   explanation type
   
3. Generate new explanations that:\\
   - Lead to the same answer as provided \\
   - Match the specified explanation type category \\
   - Sound like they come from different students at the appropriate grade level \\
   - Show variety in wording and approach while maintaining the same underlying reasoning \\
     pattern
   - Are age-appropriate in language and mathematical sophistication \\

<scratchpad>\\
Before generating the explanations, think through:\\
- What grade level is this question appropriate for?\\
- What is the specific reasoning pattern shown in the explanation type?\\
- How do the related explanations demonstrate this pattern?\\
- What variations in language and approach can I use while maintaining the same reasoning type?\\
- If it's a misconception, what is the specific mathematical error being made?\\
</scratchpad>\\

Generate \{N\} new student explanations that match the specified explanation type. Each 
explanation should be distinct but follow the same reasoning pattern. Present each explanation 
numbered and in separate tags:

<explanation\_1>
[First new student explanation]
</explanation\_1>

<explanation\_2>
[Second new student explanation]
</explanation\_2>

[Continue for all N explanations...]
\end{tcolorbox}



\onecolumn
\begin{tcolorbox}[
    colback=orange!5!white,
    colframe=orange!75!black,
    title=\textbf{Prompt: MAP-Charting dataset},
    fonttitle=\bfseries,
    breakable,
    enhanced,
    width=\textwidth,
    left=2mm,
    right=2mm,
    top=2mm,
    bottom=2mm,
    label=fig:map-charting-prompt
]
\small\ttfamily
\textbf{\#\# SYSTEM PROMPT:}

You are a mathematics education analysis expert. You need to analyze students' explanations 
for solving math problems, identify correct reasoning, vague statements, or specific 
misunderstandings, and categorize them into predefined types. Please remain professional 
and objective, focusing on the students' thought processes.\\

\textbf{\#\# CLASSIFICATION PROMPT:}

You will be analyzing a student's explanation for a math problem and classifying it into 
one of several predefined explanation types.

Here is the problem data:

<problem\_data>\\
\{PROBLEM\_DATA\}\\
</problem\_data>\\

Here are the possible student explanation types with their categories and corresponding indices:

<student\_explanation\_types>

\textbf{\#\#\# General Categories}

- [0] \textbf{Correct:NA} - The student's problem-solving approach and process are correct

- [36] \textbf{Neither:NA} - The explanation is vague, unclear, or logically incomplete

\textbf{\#\#\# Fraction Operation Misconceptions}

- [1] \textbf{Misconception:Adding\_across} - Adding numerators and denominators directly 
      (e.g., 1/3 + 2/5 = 3/8)
      
- [7] \textbf{Misconception:Denominator-only\_change} - Only denominator is changed 
      (e.g., 1/3 + 2/5 = 3/15)
      
... (6 more fraction operation types)

\textbf{\#\#\# Basic Operation Misconceptions}

- [2] \textbf{Misconception:Adding\_terms} - Using addition instead of multiplication 
      (e.g., 2y = 24 interpreted as y = 22)
      
- [8] \textbf{Misconception:Division} - Misunderstanding division concept (e.g., confusing "of" with "÷")

... (5 more basic operation types)

\textbf{\#\#\# Other Misconception Categories}

- [9] \textbf{Misconception:Duplication} - Multiplying both numerator and denominator (e.g., 2/3 × 5 = 10/15)

- [19] \textbf{Misconception:Longer\_is\_bigger} - Believing more decimal places means larger numbers

- [22] \textbf{Misconception:Not\_variable} - Not understanding variables (e.g., interpreting 2y as 2 and y)

... (18 more misconception types across various mathematical concepts)\\
</student\_explanation\_types>\\

\textbf{Your task is to determine which explanation type from STUDENT\_EXPLANATION\_TYPE best} 
\textbf{matches the Student Explanation provided in the PROBLEM\_DATA.}

The category system works as follows:

- \textbf{Correct:NA} - The student's explanation process is correct

- \textbf{Neither:NA} - The student's explanation is vague, unclear, or irrelevant to 
  the problem
  
- \textbf{Misconception:[specific\_type]} - The student's explanation contains a specific 
  misunderstanding or error in reasoning 
  (e.g., "Misconception:Incorrect\_equivalent\_fraction\_addition")

Before providing your final answer, analyze the student's explanation carefully in 
scratchpad tags. Consider:

1. What mathematical concepts or processes does the student's explanation involve?

2. Is the reasoning correct, incorrect, or unclear?

3. If incorrect, what specific type of misconception does it represent?

4. Which of the available explanation types best matches this analysis?
\\
<scratchpad>\\
{}[Your analysis here]
</scratchpad>\\

Provide the three most likely explanation type indices in order from highest to lowest 
probability. Format your answer as [idx1, idx2, idx3, ...] where idx1 is the most likely 
match, idx2 is the second most likely, and idx3 is the third most likely.
\end{tcolorbox}

\clearpage

\onecolumn
\begin{tcolorbox}[
    colback=purple!5!white,
    colframe=purple!75!black,
    title=\textbf{Prompt: Algebra Misconceptions Benchmark},
    fonttitle=\bfseries,
    breakable,
    enhanced,
    width=\textwidth,
    left=2mm,
    right=2mm,
    top=2mm,
    bottom=2mm,
    label=fig:algebra-misconception
]
\small\ttfamily
\textbf{\#\# SYSTEM PROMPT:}

You are a mathematics education diagnostic expert specializing in misconception analysis. 
Your task is to analyze students' final answers to math problems, identify whether the 
answer reflects specific mathematical misconceptions, and categorize them into predefined 
misconception types. Please remain professional and objective, focusing on the students' 
submitted answers and the underlying errors they reveal.

\textbf{\#\# CLASSIFICATION PROMPT:}

You will be analyzing a student's answer to a math problem and classifying it into one 
of several predefined misconception types.\\

Here is the problem data:

<problem\_data>\\
\{PROBLEM\_DATA\}\\
</problem\_data>\\

Here are the possible misconception types with their categories and corresponding indices:

<misconception\_types>\\
\textbf{\#\#\# Representative Misconception Categories (55 total)}\\
\textbf{\#\#\# Proportional Relationships:}

- [MaE01] when students don't understand how to represent proportional relationships

- [MaE02] Students misunderstand proportional relationships, not realizing parts must be equal

\textbf{\#\#\# Fractions:}

- [MaE06] when students inaccurately simplify fractions by guessing instead of dividing

- [MaE08] incorrectly add/subtract fractions by summing numerators and denominators separately

\textbf{\#\#\# Decimals \& Negatives:}

- [MaE16] mistakenly position decimal point left of sum, assuming units/tenths combine separately

- [MaE18] when students are unsure of correct sign when adding positive and negative numbers

\textbf{\#\#\# Ratios \& Percentages:}

- [MaE24] struggle to understand that ratios can compare same or different units

- [MaE29] incorrectly apply single proportion formula: (smaller)/(larger) = (x/100)

\textbf{\#\#\# Operations:}

- [MaE31] incorrectly assume commutative/associative properties apply to subtraction/division

- [MaE34] incorrectly perform operations left to right, neglecting order of operations

\textbf{\#\#\# Functions \& Graphing:}

- [MaE38] struggle to grasp that linear function represents consistent rate of change

- [MaE43] struggle with plotting points, reversing x- and y-coordinates

\textbf{\#\#\# Variables \& Algebra:}

- [MaE46] mistakenly perceive variables as labels/units, or associate value with alphabetical position

- [MaE51] misunderstand equal sign as "the answer is" rather than relationship between quantities

- [MaE55] struggle to recognize when to combine like terms (e.g., 4x+2x+x=7x)

... (42 more misconception types covering exponents, mixed numbers, division, 
patterns, slopes, equations, and various algebraic concepts)\\
</misconception\_types>\\

Your task is to determine which misconception type best matches the Student Answer.

Before providing your final answer, analyze carefully in scratchpad tags. Consider:

1. What is the correct answer to this problem?

2. How does the student's answer differ from the correct answer?

3. What mathematical error or misconception could lead to this specific incorrect answer?

4. Which misconception types best explain the error pattern?

5. Are there alternative misconceptions that could also explain this answer?\\

<scratchpad>\\
{}[Your analysis here]
</scratchpad>\\

Provide the three most likely misconception type indices in order from highest to lowest 
probability. Format your answer as [MaE01, MaE02, MaE03, ...] where MaE01 is the most 
likely match.
\end{tcolorbox}


\clearpage

\end{document}